\documentclass[11pt,a4paper]{article}
\usepackage{times,latexsym}
\usepackage{url}
\usepackage[T1]{fontenc}

\usepackage[utf8]{inputenc}
\usepackage{graphicx} 

\usepackage{xcolor}
\usepackage{booktabs}
\usepackage[normalem]{ulem}
\usepackage[small]{caption} 
\usepackage{amsmath}
\usepackage{interval}

\usepackage[backgroundcolor=gray!20,textsize=footnotesize]{todonotes}

\newcommand{\monoBsl}{mLM}
\newcommand{\biBsl}{LM}
\newcommand{\ourApproach}{SMaTD}
\newcommand{\langmodel}{LM}
\newcommand{\ourApproachExpanded}{\textbf{S}urrogate \textbf{Ma}chine \textbf{T}ranslation \textbf{D}etection}
\newcommand{\ourApproachCombined}{\ourApproach{}+\biBsl{}}

\usepackage{hyperref}
\usepackage{multirow}
\usepackage{makecell}
\usepackage{subcaption}
\usepackage{balance}

\usepackage[acceptedWithA]{tacl2021v1}

\usepackage{xspace,mfirstuc,tabulary}

\newif\iftaclinstructions
\taclinstructionsfalse 
\iftaclinstructions
\renewcommand{\confidential}{}
\renewcommand{\anonsubtext}{(No author info supplied here, for consistency with
TACL-submission anonymization requirements)}
\newcommand{\instr}
\fi

\iftaclpubformat

\else

\fi

\title{Automatic Machine Translation Detection Using a Surrogate Multilingual Translation Model}

\author{
   Cristian García-Romero$^\dagger$ 
   \qquad
   Miquel Esplà-Gomis$^\dagger$$^\ddagger$
   \qquad
   Felipe Sánchez-Martínez$^\dagger$$^\ddagger$
   \\
   $^\dagger$Dep. de Llenguatges i Sistemes Informàtics \\
   $^\ddagger$Institut Universitari d'Investigació Informàtica \\
   Universitat d'Alacant\\
   E-03690 Sant Vicent del Raspeig (Spain)\\
   \texttt{\{cristian.gr,miquel.espla,fsanchez\}@ua.es}
}

\begin{document}

\maketitle

\begin{abstract}
Modern machine translation (MT) systems depend on large parallel corpora, often collected from the Internet. However, recent evidence indicates that (i) a substantial portion of these texts are machine-generated translations, and (ii) an overreliance on such synthetic content in training data can significantly degrade translation quality. As a result, filtering out non-human translations is becoming an essential pre-processing step in building high-quality MT systems.
In this work, we propose a novel approach that directly exploits the internal representations of a surrogate multilingual MT model to distinguish between human and machine-translated sentences. 
Experimental results show that our method outperforms current state-of-the-art techniques, particularly for non-English language pairs, achieving gains of at least 5 percentage points of accuracy. 
\end{abstract}

\section{Introduction}

Parallel corpora are an essential resource for the development of machine translation (MT) systems. They are used both to train models from scratch and to fine-tune pre-trained systems,  particularly for domain adaptation and for enhancing performance in low-resource language pairs.

Automatic harvesting of parallel content from the Web is a common practice to build such corpora. However, some studies have highlighted a significant caveat: the Internet is increasingly populated with machine-translated content \citep{dodge2021documenting,ramirez2022human,thompson2024shocking}. Furthermore, empirical evidence shows that using large amounts of synthetic, automatically translated texts for training can degrade the 
translation performance of the resulting translation models~\citep{wu2019exploiting,jiao2021alternated}. This degradation ---which has also been observed in generative-AI models \citep{shumailov2024ai}--- underscores the importance of  distinguishing synthetic translations from human ones. 
The challenge is exacerbated by the fact that modern neural MT systems generate output that is not only grammatically fluent but also stylistically natural, making superficial quality checks insufficient.

In adjacent fields, such as the detection of AI-generated texts, a common approach is to leverage the very same large language models (LLMs) that produce the synthetic content to also detect it, following a white-box approach \citep{gehrmann2019gltr}. 
While promising, these self-referential methods rely on access to the generation model, which may not always be available.
Consequently, recent research has increasingly focused on the use of surrogate models for AI-generated text detection \citep{mitchell2023detectgpt,venkatraman2024gpt}. 
These approaches extract interpretable features ---such as log-probability distributions--- from the internal representations of a surrogate model, which are then used to classify a given text as either human- or machine-generated. 

\begin{figure}
    \centering
    \includegraphics[width=\linewidth]{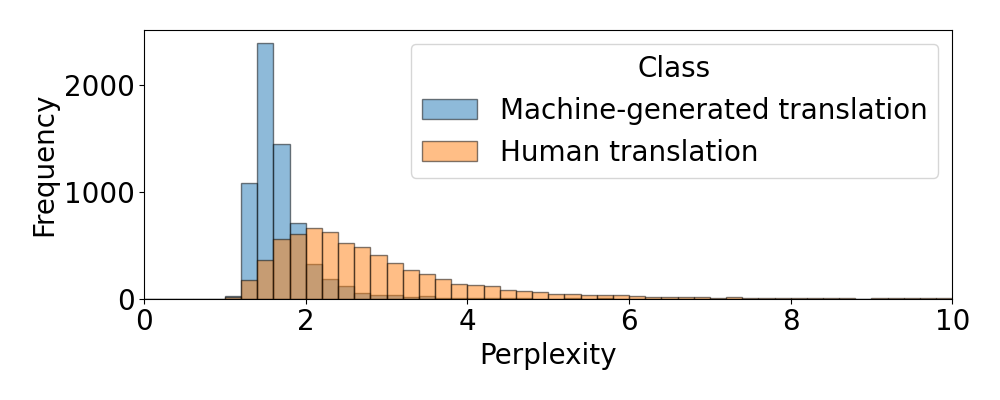}
    \caption{Per-word perplexity 
    for human and MT-generated translations (produced with MADLAD) from our English--German  training set (see Sec.~\ref{section:experimental-setting}). 
    Perplexity is obtained with NLLB 3.3B. Similar trends are observed across language pairs and MT models.}
    \label{fig:perplexity-nmt-vs-ht}
\end{figure}

Our approach builds on the latter research line but adopts a more flexible strategy: 
instead of relying on hand-crafted features derived from model internals, 
we directly use the latent representations from a pre-trained multilingual surrogate MT model ---in our experiment, NLLB~\citep{nllb}--- to detect machine-translated sentences. 
This avoids potential information loss from manual feature extraction and allows the classifier to learn relevant patterns automatically. It also enables analysis of which components of the surrogate model (e.g., specific decoder blocks) 
are more informative. 
Crucially, by decoupling detection from the generation model, our method generalizes across all language pairs supported by the surrogate. Empirical evidence ---shown in Fig.~\ref{fig:perplexity-nmt-vs-ht}--- further supports our approach: 
MT consistently shows lower per-word perplexity than human translations (HT) across a variety of MT generation models.
Our findings show that pre-trained multilingual MT models encode useful signals for 
detecting machine-translated sentences, outperforming the current state of the art based on fine-tuned pre-trained multilingual encoder models~\citep{chichirau2023automatic}.

The rest of the paper is organized as follows. The next section presents our approach in detail. Sec.~\ref{section:experimental-setting} then outlines the experimental settings, whereas  Sec.~\ref{section:results} reports and discusses the results. The final sections cover related work, conclusions, and a discussion of limitations.

\begin{figure*}[t]
    \centering
    \includegraphics[width=\linewidth]{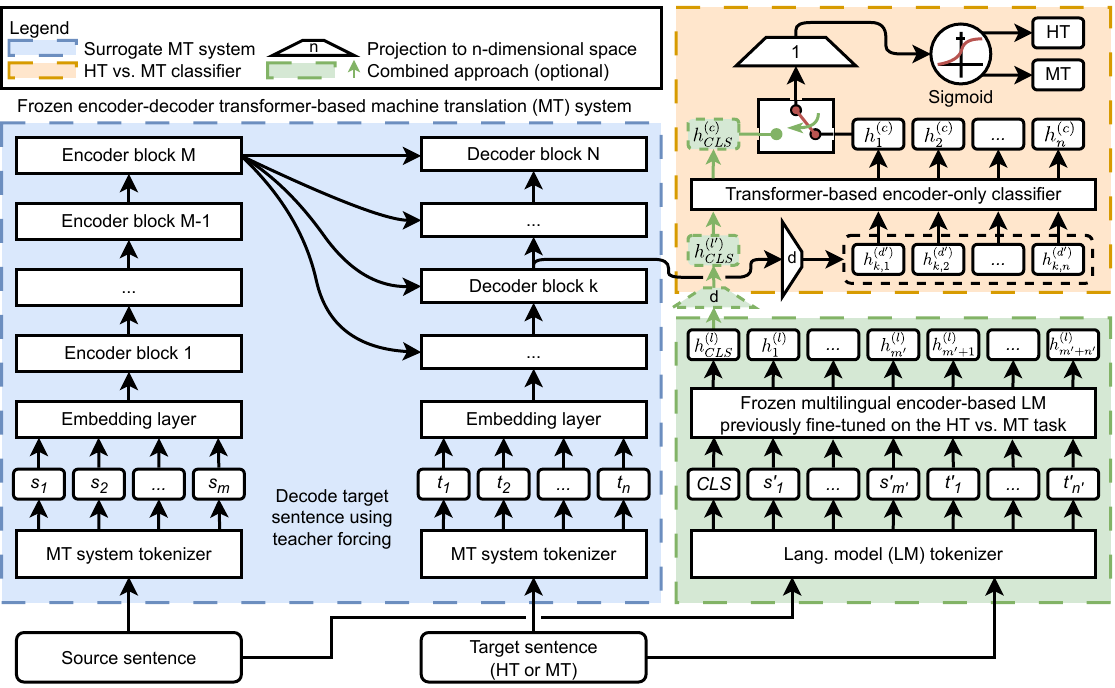}
    \caption{Architecture of \ourApproach{}/\ourApproachCombined{}. 
    Some elements (e.g., positional embeddings) have been omitted for clarity. See Sec.~\ref{section:approach} for a detailed explanation.}

    \label{fig:main-diagram}
\end{figure*}

\section{Approach}
\label{section:approach}

We formulate the HT vs.\ MT problem as a binary classification task. Given a sentence pair $(s,t)$ in languages $S$ and $T$, where $s$ is a (human-produced) source sentence and $t$ is its translation (either HT or MT), we use a pre-trained multilingual MT model to obtain representations of $t$ conditioned on $s$, which are then fed into a classifier.\footnote{A monolingual variant would consider only $t$.}

Inspired by \citet{sarvazyan2024genaios}, who address the detection of machine-generated text from LLMs, we propose a similar approach tailored to the HT vs.\ MT classification task, while avoiding the use of handcrafted features.
In our method, illustrated in Fig.~\ref{fig:main-diagram}, the tokenized source sentence 
$s_{\text{tok}}=\{s_j\}_{j=1}^m$ 
is fed into a surrogate multilingual encoder-decoder MT model 
while the tokenized translation 
$t_{\text{tok}}=\{t_i\}_{i=1}^n$ 
is provided to the decoder via teacher forcing.\footnote{Adapting the method to a multilingual decoder-only model, such as EMMA-500~\citep{ji2024emma}, would require minor adjustments.} 
We then extract the target token-level representations 
$h_{k}^{(d)}=\{h_{k,i}^{(d)}\}_{i=1}^n$
from the hidden states of a chosen decoder 
block\footnote{\emph{Blocks} refer to the components 
in a Transformer layer \citep{vaswani2017attention}: 
attention mechanism, residual connections, layer normalization, and feed-forward network. 
} $k$, where $h_{k,i}^{(d)}$ is the hidden state of the $k$-th decoder block for token $t_i$. These representations are subsequently fed into a classifier after passing it through a projection layer 
to get $h^{(d')}=\{h_{i}^{(d')}\}_{i=1}^n$. This projection layer maps the surrogate model's hidden states to the classifier's embedding space and 
decouples the classifier from the surrogate model's dimensionality (e.g., $1024$ for NLLB~1.3B vs.\ $2048$ for NLLB~3.3B). This ensures a consistent architecture across experiments, aiding interpretation and fair comparison across surrogate models of different sizes (see Sec.~\ref{ssection:surrogate-size-layer-eval}).

Our classifier is a transformer-based encoder 
with absolute positional embeddings to retain word order. It processes the output of the token-level representations $h^{(d')}$ 
generated by the projection layer 
and produces as output $h^{(c)}=\{h_i^{(c)}\}_{i=1}^n$. 
Following a BERT-style setup~\citep{devlin-etal-2019-bert}, the representation of the first token, $h_1^{(c)}$, is then passed through a feed-forward layer to a single output neuron. This output is finally normalized with a sigmoid function to estimate the probability that the target sentence is an HT. We refer to this method as \ourApproachExpanded{}, or simply \ourApproach{}.

\subsection{Combination with a Language Model}
\label{ssection:combining-approach-and-bbsl}
The classifier described above can optionally incorporate a representation from an encoder-based language model (LM) as additional input. This is done by processing a sequence of tokens consiting of a special \texttt{[CLS]}
token  followed by the tokens $\{{s'}_{j}\}_{j=1}^{m'}$ and $\{{t'}_{i}\}_{{i}=1}^{n'}$,
produced, respectively, by the LM's tokenizer for $s$ and $t$. Following a BERT-style approach \citep{devlin-etal-2019-bert}, we extract the hidden state of the first token, $h_{\text{CLS}}^{(l)}$, from the LM's final block and pass it through a projection layer to yield $h_{\text{CLS}}^{(l')}$. $h_{\text{CLS}}^{(l')}$ is then prepended to the sequence of $n$ token-level representations $h^{(d')}$ from the surrogate model to form the final input sequence for the transformer-based classifier. Finally, 
the representation of the first token produced as output by the transformer ---in this case $h_{\text{CLS}}^{(c)}$, instead of $h_1^{(c)}$--- is used for the final classification (see Fig.~\ref{fig:main-diagram}).

Preliminary experiments show that the most effective and analytically insightful configuration involves fine-tuning the  LM for HT vs.\ MT classification, and then freezing it when integrating it with the surrogate MT representations. This two-stage approach isolates the contribution of the \langmodel{} and facilitates quantifying the performance gains from combining both sources of information. We refer to this configuration as \ourApproachCombined{}.

\section{Experimental Setting}
\label{section:experimental-setting}

This section outlines the datasets, evaluation metric and baselines used for comparison, followed by training details for both \ourApproach{} and \ourApproachCombined{}.

\paragraph{Datasets.}
\label{ssection:dataset}
We build on the dataset released by \citet{chichirau2023automatic}, which leverages data from the WMT news translation shared tasks (2008--2019).\footnote{\url{https://www.statmt.org/wmt19/translation-task.html}} This dataset consists of sentence pairs where the source segments are original, and the target segments are HT. It is English-centric, covering translations into English from German (de--en), Russian (ru--en), and Chinese (zh--en). The authors supplemented the existing human translations with additional MT outputs generated using DeepL\footnote{\url{https://www.deepl.com/translator}} and Google Translate (hereafter Google).\footnote{\url{https://translate.google.com/}} 

We extend this dataset by adding new MT systems and language pairs. The evaluated systems include state-of-the-art models: MADLAD-400~\citep{kudugunta2024madlad} (multilingual; hereafter MADLAD or MADL), Opus-MT~\citep{tiedemann2020opus} (bilingual; hereafter Opus)\footnote{Although Opus is bilingual and uses separate models per language pair, we refer to it as a single system for clarity.}  and Tower~Instruct~\citep{alves2024tower} (hereafter Tower).\footnote{We use the prompt recommended by the authors: \url{https://huggingface.co/Unbabel/TowerInstruct-7B-v0.2}} Tower is an instruction-tuned, decoder-only model; its inclusion enables us to examine whether such models behave differently from traditional encoder-decoder systems in distinguishing MT from HT.

\begin{table}[tb]
\centering
\begin{small}
\begin{tabular}{ llll}
\toprule
\multicolumn{1}{c}{\thead{Lang.\\pair}} & \multicolumn{1}{c}{\thead{Training}} & \multicolumn{1}{c}{\thead{Development}} & \multicolumn{1}{c}{\thead{Test}} \\
\midrule
de--en & 08--17 & 18 & 19 \\
ru--en & 15--17 & 18 & 19 \\
en--de & 08--11,13,15--17 & 18 & 19 \\
en--ru & 08--13,15--17 & 18 & 19 \\
de--es & 08--09 & 10 & 11,13 \\
es--de & 08--09 & 10 & 11,13 \\
fi--en & - & - & 19 \\
\bottomrule
\end{tabular}
\end{small}
\caption{WMT editions used to create each data split.}
\label{table:dataset-wmt-editions}
\end{table}

The additional language pairs are: English--German (en--de), English--Russian (en--ru), 
German--Spanish (de--es), and Spanish--German (es--de). Data was collected from WMT using the same method as \citet{chichirau2023automatic}, selecting editions based on language availability (see Table~\ref{table:dataset-wmt-editions}). We use the WMT 2019 edition for testing,
the preceding one for development, and earlier editions for training, except for de--es and es--de, where the WMT 2011 and 2013 editions are used for testing 
and the WMT 2010 edition is used for development.

Unlike \citet{chichirau2023automatic}, we exclude Chinese from our experiments due to concerns about data reliability. First, zh--en exhibits unusually unstable per-token perplexity scores in our surrogate system, deviating from the consistent trends observed in Fig.~\ref{fig:perplexity-nmt-vs-ht}. Second, \citet{chichirau2023automatic} also reported that zh--en behaves differently from the rest of language pairs. Third, translations of Chinese segments in the WMT datasets consistently yield significantly lower BLEU, chrF, and xCOMET scores ---regardless of the MT system used in our experiments--- compared to the rest of language pairs.

To evaluate generalization to unseen MT systems and language pairs, we introduce M2M-100~\citep{fan2021beyond} (hereafter M2M) as a zero-shot MT system, and adopt Finnish--English (fi--en) data provided by \citet{chichirau2023automatic}, generating additional translations with our MT systems, for zero-shot language evaluation.

Each source 
sentence is
paired with its HT and with MT-generated translations from the corresponding MT systems, resulting in a balanced dataset per language and MT system. Table~\ref{table:report-dataset} shows the number of  source sentences per language pair across the training, development, and test splits.  Google, DeepL, and Tower were used for de--en 
and ru--en, while Tower, MADLAD, and Opus were used for the remaining pairs. For zero-shot evaluation, all MT systems generated fi--en translations, and M2M produced translations for all language pairs.

\begin{table}[tb]
\centering
\begin{small}
\begin{tabular}{lrrr}
\toprule
\multicolumn{1}{c}{\thead{Lang.\\pair}} & \multicolumn{1}{c}{\thead{Sentences\\training}} & \multicolumn{1}{c}{\thead{Sentences\\dev.}} & \multicolumn{1}{c}{\thead{Sentences\\test}} \\
\midrule
de--en & 8,242 & 1,498 & 2,000 \\
ru--en & 4,382 & 1,500 & 2,000 \\
en--de & 6,501 & 1,500 & 1,997 \\
en--ru & 4,935 & 1,500 & 1,997 \\
de--es &   793 &   500 & 1,101 \\
es--de &   685 &   499 & 1,104 \\
fi--en &     - &     - & 1,996 \\
\bottomrule
\end{tabular}
\end{small}
\caption{Number of source sentences per language pair in the train, dev., and test sets. Each source sentence is paired with its HT and the corresponding MT-generated translations. Counts for ru--en differ from \citet[Table~1]{chichirau2023automatic} but align with the corresponding WMT test sets.}
\label{table:report-dataset}
\end{table}

\paragraph{Evaluation.}
\label{ssection:eval-metric}
Following \citet{chichirau2023automatic} and \citet{bhardwaj2020human}, we use accuracy as our evaluation metric. F1-score could also be used, but it has been shown to strongly correlate with accuracy for HT vs.\ MT classification using pre-trained \langmodel{}s \citep{van2022automatic}.

\paragraph{Baselines.}
\label{ssection:baseline}
We compare our approach to three different baselines: LLMixtic~\citep{sarvazyan2024genaios}, a state-of-the-art method for detecting machine-generated text, and two other methods based on the current state of the art for detecting MT-generated translations~\cite{chichirau2023automatic}.

LLMixtic~\citep{sarvazyan2024genaios}, utilizes large language models (LLMs) as surrogate models and achieved state-of-the-art results in the automatic detection of monolingual machine-generated text, ranking first (out of 125 submissions) in the SemEval-2024 Task 8 \citep{wang-etal-2024-semeval-2024}. 
For our experiments, we used the authors' implementation\footnote{\url{https://github.com/jogonba2/llmixtic}} with the best-reported configuration of LLaMA-2 models~\citep{touvron2023llama2openfoundation} ---Llama-2-7b-hf, Llama-2-7b-chat-hf, Llama-2-13b-hf, and Llama-2-13b-chat-hf--- and minor modifications ---unlimited training with a patience of 6 epochs, and an inverse square root learning rate scheduler.\footnote{We used LLaMA-2 for all language pairs in our experiments because the newer LLaMA-3.1 models \citep{grattafiori2024llama3herdmodels} (Llama-3.1-8B and Llama-3.1-8B-Instruct) did not yield superior results for non-English language pairs, despite LLaMA-2 officially only supporting English.}
 
The two other baselines correspond to two versions of the state-of-the-art method proposed by \citet{chichirau2023automatic} for detecting MT-generated translations, which uses a multilingual encoder-based \langmodel. Specifically, we adopt mDeBERTaV3\textsubscript{Base}~\citep{he2023debertav3} (hereafter mDeBERTaV3) in two configurations: a \emph{monolingual} setup, where only the target sentence is used as input, and a \emph{bilingual} setup, where both the source and target sentences are provided. The latter matches the configuration used for the \langmodel{} in \ourApproachCombined{}. As with \ourApproachCombined{}, we use the 
first token (\texttt{[CLS]}) embedding to estimate the probability that the target sentence is HT or MT.
We follow the best configuration from \citet{chichirau2023automatic}, with minor adjustments: unlimited training with a patience of 6 epochs, $10$\% dropout (classifier, attention, and feed-forward layers), learning rate of $10^{-5}$, and inverse square root learning rate scheduler with $400$ warm-up steps. We use the AdamW optimizer \citep{loshchilov2017decoupled} with default settings and no weight decay. 
All reported results 
were obtained using our reimplementation, which successfully reproduced the original results.

\paragraph{\ourApproach{} training.}
\label{ssection:model-training}
For \ourApproach{}, we use a standard encoder-only Transformer \citep{vaswani2017attention} as the binary classifier, trained from scratch, and NLLB-200~\citep{nllb} as the surrogate multilingual translation model.

We adopt the same training setup as the \citet{chichirau2023automatic}'s baselines, but with a learning rate of $10^{-4}$ and $10$\% dropout on positional embeddings. Based on preliminary experiments we use $3$ encoder layers, $4$ attention heads, a $2048$-dimensional feed-forward layer, and a model dimensionality of $512$. The parameters of the surrogate model remain frozen during 
training.

\paragraph{\ourApproachCombined{} training.}
\label{ssection:combined-approach}
We follow the two-stages training procedure described in Sec.~\ref{ssection:combining-approach-and-bbsl}, using the same mDeBERTaV3 language model from the bilingual baseline, which is pre-trained on the HT vs.\ MT classification task.
To prevent the classifier from relying solely on mDeBERTaV3 and ignore the surrogate model, we apply \emph{stochastic depth}~\citep{huang2016deep}, 
a form of dropout at the representation level. Based on preliminary experiments, we set its probability to $0.7$.
All other hyperparameters match those used for training mDeBERTaV3 (baselines) and \ourApproach{}.

\section{Results and Discussion}
\label{section:results}
We begin by analyzing how the size of the surrogate MT model and the choice of decoder block impact classification performance (Sec.~\ref{ssection:surrogate-size-layer-eval}). We then evaluate \ourApproach{} and \ourApproachCombined{} 
(Sec.~\ref{ssection:main-results}), examining their accuracy, their generalization to unseen languages and MT systems 
(Sec.~\ref{ssection:cross-evaluation-results}), and their transferability when trained on multilingual or multi-MT datasets (Sec.~\ref{ssection:transferability-assessment}).

All reported accuracy scores, computed over the test set, correspond to models selected based on their performance on the development set. Specifically, for each approach, we selected, out of three independent training executions, the model checkpoint that achieved the highest development set accuracy within its run (using early stopping with exhaustion of patience). See Appendix~\ref{section:appendix-run-variability} for details on the variation over training executions. 
Statistically significant differences are  reported in all tables of this section, as computed via approximate randomization (10,000 iterations; $p < 0.05$).

\subsection{Surrogate MT Model Size and Layer Evaluation}
\label{ssection:surrogate-size-layer-eval}

\begin{figure}[t]
\centering
\includegraphics[width=\linewidth]{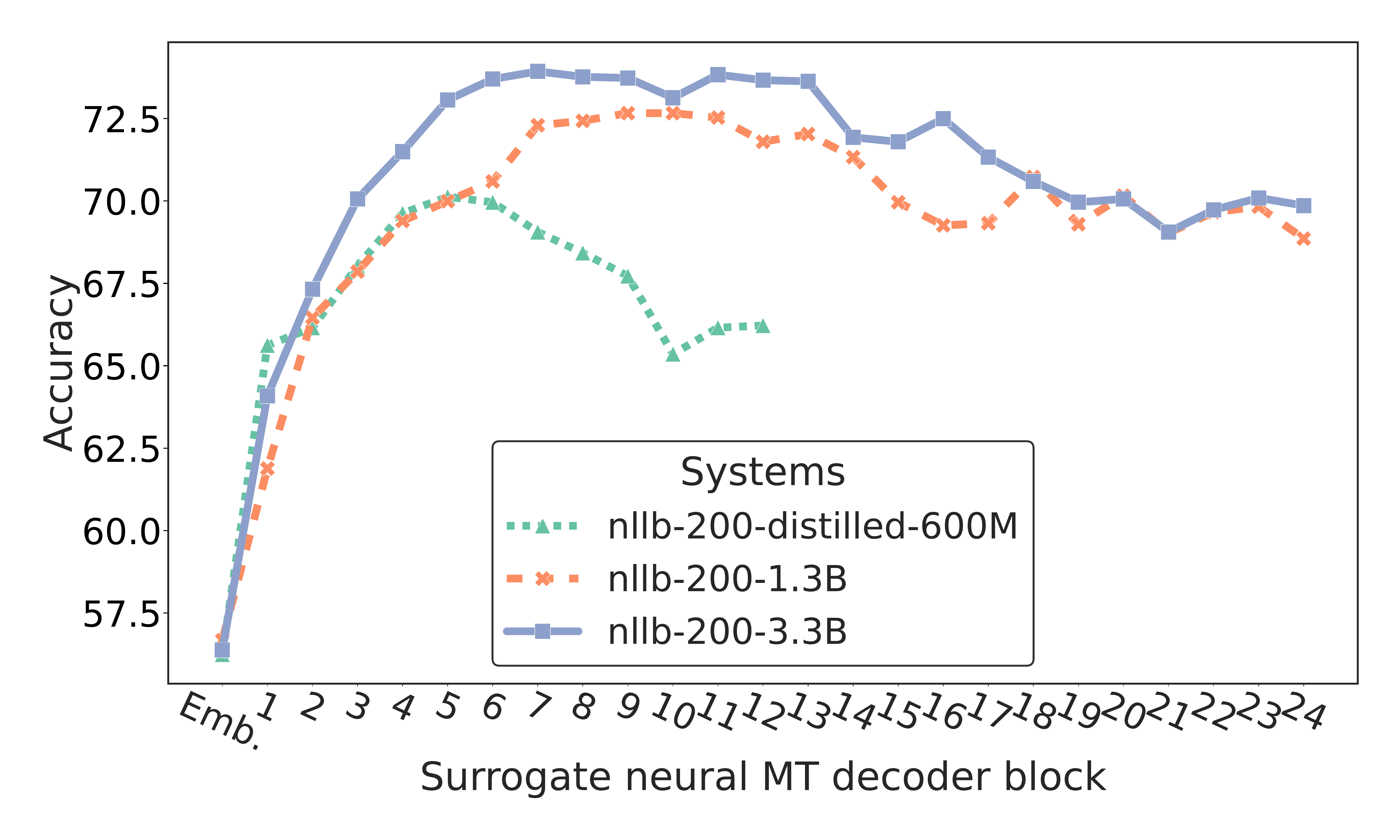}
\caption{Accuracy on the development set
for DeepL (de--en) evaluated using 
three different sizes of the NLLB surrogate 
MT model. 
For each surrogate model, we evaluate the use of the hidden state of different decoder blocks as input to the classifier; note that the 600M model has fewer decoder blocks by design.}
\label{fig:nllb-size-and-layers}
\end{figure}

Fig.~\ref{fig:nllb-size-and-layers} shows the accuracy for de--en with DeepL, comparing three NLLB surrogate model sizes (600M, 1.3B, 3.3B) across all decoder blocks. 
The results show that the 3.3B model consistently outperforms the others models, with 1.3B outperforming 600M at every decoder block. Similar patterns were observed for other language pairs and MT systems. Based on these observations, we report results exclusively for the 3.3B model in subsequent experiments.

Interestingly, Fig.~\ref{fig:nllb-size-and-layers} shows a consistent pattern: both the first and last decoder blocks underperform compared to middle blocks. 
To identify the best-performing block, we aggregate the accuracy across all language pairs and MT systems on the development set: 
the 
lowest decoder block (embedding layer) yields the weakest performance, with an accuracy of $58.70$\%, while 
the 10th block achieves the highest accuracy 
($72.92$\%).\footnote{Across decoder blocks, aggregated performance has a median of $69.87$\%, a mean of $68.53$\% and a standard deviation of $4.05$.} 
To assess the sensitivity of decoder block choice, we compute the difference in accuracy between the 10th block and the best-performing block in each experiment. 
The results indicate that performance is highly robust to the block choice  (median: $-0.68$; mean: $-0.98$; std. dev.: $1.23$). Accordingly, all subsequent experiments where performed using the 10th decoder block.\footnote{Note that the 10th layer may not be optimal in the cross-evaluation experiments reported in Sec.~\ref{ssection:cross-evaluation-results}.}

\subsection{Main Results}
\label{ssection:main-results}

\begin{table}[t!]
    \centering
    \begin{small}
    \begin{tabular}{@{\hspace{1ex}}c@{\hspace{2ex}}l|@{\hspace{1ex}}rrr@{\hspace{1ex}}}
    \toprule
    \multirow{3}{*}{\makecell{Train.\\\& eval.\\lang.}} & \multirow{3}{*}{\thead{Approach}} & \multicolumn{3}{c}{\multirow{2}{*}{\thead{Train. \& eval. MT system}}} \\

    & & \multicolumn{3}{c}{} \\
    
    & & Google& DeepL& Tower \\

    \midrule
    \multirow{5}{*}{de--en}

    & LLMixtic & 49.53 & 49.98 & 56.70 \\
    &  \monoBsl{}  &  \underline{63.95} &  \underline{63.10} &  \underline{61.45}  \\
    &  \biBsl{}  &  \underline{75.52} &  \underline{72.53} &  \underline{70.38}  \\
    &  \ourApproach{}  &  75.88 &  72.90 &  71.32  \\
    &   \ourApproachCombined{} &  $\dagger$\underline{\textbf{77.28}} &  $\dagger$\underline{\textbf{75.25}} &  $\dagger$\textbf{71.80}  \\

    \midrule
    \multirow{5}{*}{ru--en}

    & LLMixtic & 49.23 & 50.38 & 49.25 \\
    &  \monoBsl{}  &  \underline{59.12} &  \underline{59.58} &  \underline{56.60}  \\
    &  \biBsl{}  &  \underline{70.32} &  \underline{64.05} &  \underline{64.07}  \\
    &  \ourApproach{}  &  \underline{\textbf{74.40}} &  \textbf{65.40} &  \underline{67.40}  \\
    &   \ourApproachCombined{} &  $\dagger$74.08 &  $\dagger$65.10 &  $\dagger$\textbf{67.42}  \\

    \midrule

    & & MADL.& Opus& Tower \\

    \midrule
    \multirow{5}{*}{en--de}
    
    & LLMixtic & 52.35 & 50.48 & 53.93 \\
    &  \monoBsl{}   &  \underline{59.49} &  \underline{65.20} &  \underline{56.08} \\
    &  \biBsl{}   &  \underline{70.76} &  \underline{74.39} &  \underline{60.72} \\
    &  \ourApproach{}   &  \underline{72.26} &  \underline{76.09} &  60.14 \\
    &   \ourApproachCombined{}  &  $\dagger$\underline{\textbf{74.21}} &  $\dagger$\underline{\textbf{78.54}} & $\dagger$\underline{\textbf{62.27}} \\

    \midrule
    \multirow{5}{*}{en--ru}

    & LLMixtic & 51.18 & 53.98 & 51.83 \\
    &  \monoBsl{}  &  \underline{63.52} &  \underline{68.00} &  \underline{58.76} \\
    &  \biBsl{}   &  \underline{68.70} &  \underline{70.38} &  \underline{60.79} \\
    &  \ourApproach{}   &  \underline{73.99} &  71.48 &  \underline{64.17} \\
    &   \ourApproachCombined{}  &  $\dagger$\textbf{74.06} &  $\dagger$\underline{\textbf{73.41}} &  $\dagger$\textbf{64.62} \\

    \midrule
    \multirow{5}{*}{de--es}

    & LLMixtic & 55.86 & 51.50 & 55.86 \\
    &  \monoBsl{}   &  58.31 &  \underline{66.26} &  \underline{60.63} \\
    &  \biBsl{}   &  \underline{66.35} &  66.39 &  \underline{63.49} \\
    &  \ourApproach{}   &  \underline{\textbf{80.20}} & \underline{\textbf{75.11}} &  \underline{\textbf{71.71}} \\
    &   \ourApproachCombined{}  &  $\dagger$\underline{77.34} &  $\dagger$73.66 &  $\dagger$70.48 \\

    \midrule
    \multirow{5}{*}{es--de}

    & LLMixtic & 55.53 & 50.41 & 50.82 \\
    &  \monoBsl{}   &  \underline{59.10} &  \underline{62.64} &  \underline{55.30} \\
    &  \biBsl{}   &  \underline{67.30} &  63.68 &  \underline{61.68} \\
    &  \ourApproach{}   &  \underline{76.95} &  \underline{\textbf{77.31}} &  \underline{\textbf{72.46}} \\
    &   \ourApproachCombined{}  &  $\dagger$\textbf{78.03} &  $\dagger$\underline{73.55} &  $\dagger$\underline{64.31} \\

    \bottomrule
    \end{tabular}
    \end{small}
    \caption{
    Accuracy obtained 
    by the five approaches under evaluation.
    Best scores are in bold. Scores whose 
    difference from 
    the score immediately above is statistically significant 
    are underlined. 
    $\dagger$ indicates statistically significant differences between  \ourApproachCombined{} and \biBsl{}.
    }
    \label{table:results-baseline-vs-ours-vs-frozen}
\end{table}

Table~\ref{table:results-baseline-vs-ours-vs-frozen} 
shows the results for 
the baseline approaches ---monolingual LLMixtic and language model (\monoBsl{}), and the bilingual language model (\biBsl{})--- together with
\ourApproach{}, and \ourApproachCombined{}, the combination of the last two (see Sec.~\ref{ssection:combining-approach-and-bbsl}). 

The results show a clear trend: either \ourApproachCombined{}  or \ourApproach{} consistently outperform 
all baselines. In particular, \ourApproach{} yields results comparable to the bilingual baseline (within $\pm1$ accuracy point) in certain cases 
(e.g., de--en for all MT systems). 
However, in most cases, 
\ourApproach{} achieves statistically-significant improvements of at least one accuracy point. Remarkably, \ourApproachCombined{} outperforms the LM baseline by a  statistically-significant margin in all cases.

Notably, the results for non-English language pairs are particularly strong, with gains of at least 5 accuracy points and even 10 
in some cases (de--es for MADLAD, es--de for Opus and Tower). Since the bilingual baseline model is English-centric ---it has been trained on CC-100~\citep[Fig.~1]{conneau2020unsupervised}, where English is the most represented language--- we argue that it might face greater challenges with non-English language pairs. In contrast, our surrogate model 
has been trained on multiple language directions beyond English-centric pairs \citep[Sec.~8.1]{nllb}. This difference may explain, at least partially, why \ourApproach{} achieves such significant accuracy improvements. 
In line with this observation, 
\ourApproachCombined{} generally outperforms 
\ourApproach{} when applied to language pairs including English, while \ourApproach{} 
performs best in the rest of cases.

\begin{table*}[t!]
    \centering

    \begin{subtable}[t]{0.48\textwidth}
    \centering
    \begin{small}
    \begin{tabular}{@{\hspace{1ex}}c@{\hspace{1em}}l@{\hspace{1ex}}|@{\hspace{1ex}}r@{\hspace{1ex}}r@{\hspace{1ex}}r@{\hspace{1ex}}||@{\hspace{1ex}}r@{\hspace{1ex}}}
    \toprule
    \multirow{3}{*}{\makecell{Train.\\\& test\\lang.}} & \multirow{3}{*}{\makecell{Train.\\on MT\\system}} & \multicolumn{4}{c}{\multirow{2}{*}{\thead{Test \biBsl{} on MT system}}} \\
    & & \multicolumn{4}{c}{} \\

    & & Google& DeepL& Tower& M2M \\

    \midrule
    \multirow{3}{*}{de--en}
    &  Google &  \textbf{75.52} &  66.68 &  66.35 &  \textbf{67.20} \\
    &  DeepL &  72.85 &  \textbf{72.53} &  68.83 &  64.60 \\
    &  Tower &  70.42 &  67.30 &  \textbf{70.38} &  65.43 \\

    \midrule
    \multirow{3}{*}{ru--en}
    &  Google &  \textbf{70.32} &  58.32 &  61.68 &  \textbf{66.70} \\
    &  DeepL &  63.40 &  \textbf{64.05} &  57.73 &  59.35 \\
    &  Tower &  65.72 &  58.10 &  \textbf{64.07} &  64.17 \\

    \midrule
    & & MADL.& Opus& Tower& M2M \\

    \midrule
    \multirow{3}{*}{en--de}
    &  MADL. &  \textbf{70.76} &  68.23 & \textbf{61.67} & 68.90 \\
    &  Opus &  68.03 &  \textbf{74.39} &  \underline{59.44} &  \textbf{73.31} \\
    &  Tower &  54.98 &  55.53 &  60.72 &  57.04 \\

    \midrule
    \multirow{3}{*}{en--ru}
    &  MADL. &  \textbf{68.70} &  61.64 &  58.19 &  70.56 \\
    &  Opus &  65.70 &  \textbf{70.38} &  56.61 &  \textbf{71.58} \\
    &  Tower &  61.97 &  57.74 &  \textbf{60.79} &  64.62 \\

    \midrule
    \multirow{3}{*}{de--es}
    &  MADL. &  66.35 &  61.76 &  63.67 &  64.35 \\
    &  Opus &  \textbf{67.57} &  \textbf{66.39} &  \textbf{64.31} &  \textbf{67.17} \\
    &  Tower &  64.08 &  58.08 &  63.49 &  61.85 \\

    \midrule
    \multirow{3}{*}{es--de}
    &  MADL. &  \textbf{67.30} &  \textbf{66.08} &  \textbf{62.91} &  \textbf{68.39} \\
    &  Opus &  63.45 &  63.68 &  60.24 &  64.58 \\
    &  Tower &  63.09 &  62.36 &  61.68 &  63.81 \\

    \bottomrule
    \end{tabular}
    \end{small}
    \end{subtable}
    \hfill
    \begin{subtable}[t]{0.48\textwidth}
    \centering
    \begin{small}
    \begin{tabular}{@{\hspace{1ex}}c@{\hspace{1em}}l@{\hspace{1ex}}|@{\hspace{1ex}}r@{\hspace{1ex}}r@{\hspace{1ex}}r@{\hspace{1ex}}||@{\hspace{1ex}}r@{\hspace{1ex}}}
    \toprule
    \multirow{3}{*}{\makecell{Train.\\\& test\\lang.}} & \multirow{3}{*}{\makecell{Train.\\on MT\\system}} & \multicolumn{4}{c}{\multirow{2}{*}{\thead{Test \ourApproach{} on MT system}}} \\
    & & \multicolumn{4}{c}{} \\

    & & Google& DeepL& Tower& M2M \\

    \midrule
    \multirow{3}{*}{de--en}
    &  Google &  \textbf{75.88} &  \underline{70.05} &  \underline{69.60} &  \underline{\textbf{70.80}} \\
    &  DeepL &  74.30 &  \textbf{72.90} &  70.10 &  \underline{68.25} \\
    &  Tower &  \underline{73.58} &  \underline{70.60} &  \textbf{71.32} &  \underline{68.20} \\

    \midrule
    \multirow{3}{*}{ru--en}
    &  Google &  \underline{\textbf{74.40}} &  \underline{61.43} &  \underline{64.47} &  \underline{\textbf{72.03}} \\
    &  DeepL &  \underline{67.35} &  \textbf{65.40} &  \underline{60.37} &  \underline{63.30} \\
    &  Tower &  \underline{73.10} &  \underline{61.98} &  \underline{\textbf{67.40}} &  \underline{70.67} \\

    \midrule
    & & MADL.& Opus& Tower& M2M \\

    \midrule
    \multirow{3}{*}{en--de}
    &  MADL. &  \underline{\textbf{72.26}} &  \underline{71.58} &  \textbf{60.82} &  \underline{73.36} \\
    &  Opus &  68.48 &  \underline{\textbf{76.09}} &  57.51 &  \underline{\textbf{75.19}} \\
    &  Tower &  \underline{59.81} &  \underline{62.57} &  60.14 &  \underline{65.47} \\

    \midrule
    \multirow{3}{*}{en--ru}
    &  MADL. &  \underline{\textbf{73.99}} &  \underline{68.43} &  \underline{62.24} &  \underline{\textbf{76.16}} \\
    &  Opus &  \underline{70.51} &  \textbf{71.48} &  \underline{60.94} &  \underline{73.61} \\
    &  Tower &  \underline{67.78} &  \underline{65.35} &  \underline{\textbf{64.17}} &  \underline{69.00} \\

    \midrule
    \multirow{3}{*}{de--es}
    &  MADL. &  \underline{\textbf{80.20}} &  \underline{72.66} &  64.35 &  \underline{77.34} \\
    &  Opus &  \underline{78.25} &  \underline{\textbf{75.11}} &  65.49 &  \underline{\textbf{77.70}} \\
    &  Tower &  \underline{74.21} &  \underline{67.39} &  \underline{\textbf{71.71}} &  \underline{73.25} \\

    \midrule
    \multirow{3}{*}{es--de}
    &  MADL. &  \underline{\textbf{76.95}} &  \underline{76.40} &  63.13 &  \underline{79.12} \\
    &  Opus &  \underline{75.95} &  \underline{\textbf{77.31}} &  60.73 &  \underline{\textbf{79.62}} \\
    &  Tower &  \underline{76.49} &  \underline{74.86} &  \underline{\textbf{72.46}} &  \underline{78.99} \\

    \bottomrule
    \end{tabular}
    \end{small}
    \end{subtable}

    \caption{Accuracy obtained by the bilingual baseline (LM; left table) and \ourApproach{} (right table) in the cross-MT setting.  Rows correspond to the MT systems for which models were trained, and columns to the MT systems on which models were evaluated. The M2M model is only used for testing (zero-shot). Best scores are in bold. Underlined values indicate statistically significant superior performance across systems. 
    }
    \label{table:cross-eval-mt-single-models}
\end{table*}

Contrary to expectations, the monolingual baseline LLMixtic performed the worst of all systems, even underperforming \monoBsl{}. This poor result persisted despite exploring alternative configurations (e.g., LLaMA-3.1; see Sec.~\ref{section:experimental-setting}). Although LLMixtic excelled at SemEval's English AI-text detection task ---a result we were able to successfully replicate with the authors' implementation--- 
it failed to converge on our MT detection task. While a complete analysis is beyond our scope, this failure may be attributed to significant distributional differences between MT and general AI-generated text, to LLMixtic requiring substantially more data than our dataset provided (though the dataset sufficed for the other models), or to the SemEval dataset being comparatively easier to classify.

In the following sections, only the 
bilingual baseline is used as a reference for two reasons: (i) 
it outperforms 
both monolingual baselines (LLMixtic and \monoBsl{}) in all cases, and (ii) like \ourApproach{}, it leverages both source and target texts, enabling a fairer comparison.

\subsection{Cross-Evaluation}
\label{ssection:cross-evaluation-results}

We analyze the generalization capabilities of the bilingual baseline (\biBsl{}) and \ourApproach{} when applied to languages and MT systems not seen during training in  cross-MT and cross-lingual settings.

\begin{table*}[t!]
   \centering
   \begin{small}
   \begin{tabular}{ccc|rrrrrr||r}
   \toprule
   \multirow{2}{*}{\makecell{Train. \& test\\MT}} & \multirow{2}{*}{\thead{Approach}} & \multirow{2}{*}{\makecell{Train. lang.}} & \multicolumn{7}{c}{\thead{Test on lang.}} \\
   
   & & & de--en& ru--en& en--de& en--ru& de--es& es--de& fi--en \\

   \midrule
   \multirow{4}{*}{Google}
   &  \multirow{2}{*}{\biBsl{}}
   &  de--en &  \textbf{75.52} &  68.35 &  - &  - &  - &  - &  \textbf{73.90} \\
   &    &  ru--en &  67.77 &  \textbf{70.32} &  - &  - &  - &  - &  71.99 \\

   \cline{3-10}\\[-4mm]
   &  \multirow{2}{*}{\ourApproach{}}
   &  de--en &  \textbf{75.88} &  \underline{\textbf{74.50}} &  - &  - &  - &  - &  \underline{\textbf{77.18}} \\
   &    &  ru--en &  \underline{69.85} &  \underline{74.40} &  - &  - &  - &  - &  72.04 \\

   \midrule
   \multirow{4}{*}{DeepL}
   &  \multirow{2}{*}{\biBsl{}}
   &  de--en &  \textbf{72.53} &  59.00 &  - &  - &  - &  - &  66.01 \\
   &    &  ru--en &  62.37 &  \textbf{64.05} &  - &  - &  - &  - &  \textbf{66.31} \\

   \cline{3-10}\\[-4mm]
   &  \multirow{2}{*}{\ourApproach{}}
   &  de--en &  \textbf{72.90} &  \underline{63.67} &  - &  - &  - &  - &  \underline{\textbf{70.54}} \\
   &    &  ru--en &  63.42 &  \textbf{65.40} &  - &  - &  - &  - &  66.38 \\

   \midrule
   \multirow{12}{*}{Tower}
   &  \multirow{6}{*}{\biBsl{}}
   &  de--en &  \textbf{70.38} &  \textbf{67.37} &  58.69 &  55.13 &  58.36 &  56.20 &  \underline{\textbf{66.28}} \\
   &    &  ru--en &  63.85 &  64.07 &  55.68 &  52.35 &  55.31 &  55.48 &  \underline{64.18} \\
   &    &  en--de &  \underline{59.13} &  \underline{58.10} &  \textbf{60.72} &  59.49 &  61.13 &  \textbf{66.44} &  \underline{51.63} \\
   &    &  en--ru &  56.85 &  57.95 &  57.79 &  \textbf{60.79} &  58.95 &  61.19 &  51.95 \\
   &    &  de--es &  \underline{59.97} &  \underline{61.83} &  52.58 &  52.68 & \textbf{63.49} &  57.20 &  \underline{56.39} \\
   &    &  es--de &  58.13 &  57.67 &  57.96 &  54.86 &  60.35 &  61.68 &  51.83 \\

   \cline{3-10}\\[-4mm]
   &  \multirow{6}{*}{\ourApproach{}}
   &  de--en &  \textbf{71.32} &  \underline{\textbf{69.82}} &  57.39 &  \underline{61.27} &  \underline{62.85} &  57.07 &  \textbf{61.87} \\
   &    &  ru--en &  65.47 &  \underline{67.40} &  \underline{57.84} &  \underline{56.94} &  \underline{60.35} &  \underline{58.92} &  58.64 \\
   &    &  en--de &  56.48 &  55.58 &  \textbf{60.14} &  60.49 &  \underline{67.26} &  \underline{68.34} &  48.27 \\
   &    &  en--ru &  \underline{61.87} &  \underline{63.38} &  \underline{59.91} &  \underline{\textbf{64.17}} &  \underline{68.17} &  \underline{67.48} &  \underline{54.66} \\
   &    &  de--es &  51.80 &  53.03 &  53.73 &  53.58 &  \underline{\textbf{71.71}} &  59.10 &  46.39 \\
   &    &  es--de &  \underline{60.43} &  \underline{60.40} &  58.21 &  \underline{60.12} &  \underline{68.35} &  \underline{\textbf{72.46}} &  50.63 \\

   \midrule
   \multirow{8}{*}{MADL.}
   &  \multirow{4}{*}{\biBsl{}}
   &  en--de &  - &  - &  \textbf{70.76} &  66.40 &  55.81 &  64.54 &  56.99 \\
   &    &  en--ru &  - &  - &  61.57 &  \textbf{68.70} &  56.36 &  59.60 &  55.84 \\
   &    &  de--es &  - &  - &  53.43 &  55.51 &  \textbf{66.35} &  59.78 &  58.44 \\
   &    &  es--de &  - &  - &  62.49 &  59.59 &  64.35 &  \textbf{67.30} &  \textbf{58.72} \\

   \cline{3-10}\\[-4mm]
   &  \multirow{4}{*}{\ourApproach{}}
   &  en--de &  - &  - &  \underline{\textbf{72.26}} &  \underline{72.53} &  \underline{77.61} &  \underline{74.91} &  \underline{67.76} \\
   &    &  en--ru &  - &  - &  \underline{66.35} &  \underline{\textbf{73.99}} &  \underline{77.70} &  \underline{64.04} &  \underline{67.08} \\
   &    &  de--es &  - &  - &  \underline{63.40} &  \underline{69.48} &  \underline{\textbf{80.20}} &  60.91 &  \underline{66.78} \\
   &    &  es--de &  - &  - &  \underline{67.88} &  \underline{68.45} &  \underline{71.80} &  \underline{\textbf{76.95}} &  \underline{\textbf{70.74}} \\

   \midrule

   \multirow{8}{*}{Opus}
   &  \multirow{4}{*}{\biBsl{}}
   &  en--de &  - &  - &  \textbf{74.39} &  \underline{66.07} &  55.04 &  64.18 &  57.04 \\
   &    &  en--ru &  - &  - &  65.75 &  \textbf{70.38} &  57.18 &  \textbf{64.67} &  54.01 \\
   &    &  de--es &  - &  - &  54.53 &  54.01 &  \textbf{66.39} &  59.47 &  \textbf{58.37} \\
   &    &  es--de &  - &  - &  62.59 &  56.79 &  60.54 &  63.68 &  56.91 \\

   \cline{3-10}\\[-4mm]
   &  \multirow{4}{*}{\ourApproach{}}
   &  en--de &  - &  - &  \underline{\textbf{76.09}} &  63.75 &  \underline{69.35} &  63.22 &  \underline{64.28} \\
   &    &  en--ru &  - &  - &  67.23 &  \textbf{71.48} &  \underline{73.39} &  \underline{74.64} &  \underline{60.25} \\
   &    &  de--es &  - &  - &  \underline{71.93} &  \underline{65.52} &  \underline{\textbf{75.11}} &  \underline{69.47} &  \underline{\textbf{67.16}} \\
   &    &  es--de &  - &  - &  \underline{71.91} &  \underline{60.64} &  \underline{75.07} &  \underline{\textbf{77.31}} &  \underline{65.01} \\

   \bottomrule
   \end{tabular}
   \end{small}
   \caption{Accuracy for the bilingual baseline (\biBsl{}) and \ourApproach{} in the cross-lingual setting.
   Rows correspond to the language pairs for which models were trained, and columns to the language pairs on which models were evaluated.
   The language pair fi--en is only used for testing (zero-shot). Best scores are in bold. Underlined values indicate statistically significant superior performance across systems.
   }
   \label{table:cross-eval-lang-single-models}
\end{table*}

\paragraph{Cross-MT.} Table~\ref{table:cross-eval-mt-single-models} shows the results for the cross-MT evaluation. As expected, 
the best results are located along the diagonal (i.e., 
when the same MT system is used for training and testing).\footnote{Diagonal results correspond to rows of Table~\ref{table:results-baseline-vs-ours-vs-frozen}.} 
Furthermore, \ourApproach{} consistently matches or surpasses the baseline when the MT systems used for training and testing are different.
The de--en case is particularly interesting: while diagonal results for \ourApproach{} are similar to the baseline, most off-diagonal values are significantly higher, suggesting strong generalization. Notable improvements are also observed in the zero-shot setting (M2M): \ourApproach{} achieves statistically significant gains in accuracy 
in all 
cases, especially for de--es and es--de, with gains of 10--15 accuracy points, compared to diagonal differences of 5--10 points.

\paragraph{Cross-Lingual.} Table~\ref{table:cross-eval-lang-single-models} presents the results of the cross-lingual evaluation. As for the cross-MT evaluation,
models trained on a particular language generally perform best when evaluated on the same language.\footnote{Diagonal results correspond to columns of Table~\ref{table:results-baseline-vs-ours-vs-frozen}.} Regarding off-diagonal results, \ourApproach{} yields stronger generalization. Notably, the results for DeepL improve cross-lingual accuracy by up to 6 accuracy points, despite similar performance on the diagonal. 
Tower shows some inconsistency: \ourApproach{} achieves better cross-lingual results across all MT systems when evaluating en--ru, de--es, es--de and en--de, but results are mixed for the rest of language pairs; 
nevertheless, most improvements are statistically significant. 
For Opus and MADLAD, \ourApproach{} achieves statistically-significant higher accuracies in most cases. 
In the zero-shot setting (fi--en), \ourApproach{} consistently outperforms the baseline, particularly when models are trained on de--en, or de--es and es--de when de--en is not present (i.e., MADLAD and Opus).

\subsection{Transferability Assessment}
\label{ssection:transferability-assessment}

This section explores the capabilities of \ourApproach{} and the bilingual baseline (\biBsl{}) to transfer knowledge across languages and/or MT systems.
To do so, we create two new datasets:
(i) a multi-MT dataset that compiles, for each language pair, all MT samples in our initial dataset (see Sec.~\ref{ssection:dataset}); and
(ii) a multilingual dataset that combines, for each MT system, data from all available language pairs. 

\begin{table*}[tb]
    \centering
    \begin{subtable}[t]{0.48\textwidth}
    \centering
    \begin{small}
    \begin{tabular}{c|rrr||r}
    \toprule
    \multirow{2}{*}{\makecell{Train. \&\\test lang.}} & \multicolumn{4}{c}{\thead{Test of \biBsl{} on MT system}} \\
    & Google& DeepL& Tower& M2M \\

    \midrule
    de--en &  71.68          & 69.42          &  68.65          &  65.82 \\
    ru--en &  65.97          & 60.50          &  61.52          &  63.05 \\

    \midrule

    & MADL.& Opus& Tower& M2M \\

    \midrule
    en--de &  \underline{\textbf{68.58}} &  \underline{\textbf{70.16}} &  \textbf{61.79} &  \textbf{70.73} \\
    en--ru &  69.45          &  65.52          &  58.36          &  73.39 \\
    de--es &  62.62          &  60.31          &  60.76          &  62.08 \\
    es--de &  60.73          &  60.64          &  59.33          &  60.82 \\

    \bottomrule
    \end{tabular}
    \end{small}
    \end{subtable}
    \hfill
    \begin{subtable}[t]{0.48\textwidth}
    \centering
    \begin{small}
    
    \begin{tabular}{c|rrr||r}
    \toprule
    \multirow{2}{*}{\makecell{Train. \&\\test lang.}} & \multicolumn{4}{c}{\thead{Test of \ourApproach{} on MT system}} \\
    & Google& DeepL& Tower& M2M \\

    \midrule
    de--en &  \underline{\textbf{75.35}} &  \underline{\textbf{72.43}} &  \underline{\textbf{70.45}} &  \underline{\textbf{69.65}} \\
    ru--en &  \underline{\textbf{70.28}} &  \underline{\textbf{64.15}} &  \underline{\textbf{64.90}} &  \underline{\textbf{66.43}} \\

    \midrule

    & MADL.& Opus& Tower& M2M \\

    \midrule
    en--de &  66.07          &  67.73          &  60.59          &  69.50 \\
    en--ru &  \underline{\textbf{71.98}} &  \underline{\textbf{68.80}} &  \underline{\textbf{61.19}} &  \underline{\textbf{75.59}} \\
    de--es &  \underline{\textbf{78.25}} &  \underline{\textbf{74.02}} &  \underline{\textbf{67.35}} &  \underline{\textbf{76.79}} \\
    es--de &  \underline{\textbf{75.91}} &  \underline{\textbf{76.18}} &  \underline{\textbf{71.38}} &  \underline{\textbf{78.53}} \\

    \bottomrule
    \end{tabular}
    \end{small}
    \end{subtable}

    \caption{Accuracy obtained by the bilingual baseline (LM; left table) and \ourApproach{} (right table) in the multi-MT setting.     
    Results are provided per MT system (columns)
    for models trained on all the available MT systems for each language pair (rows). 
    M2M is only used for testing (zero-shot). 
    Best scores across approaches are in bold. Underlined values indicate statistically significant superior performance across systems. 
    }
    \label{table:transferability-multiple-mt}
\end{table*}

\paragraph{Class Balance in New Datasets.}
The multi-MT dataset is unbalanced in terms of the HT/MT class distribution: for each HT sample, there are three MT samples, each generated by a different MT system. 
To mitigate this during training, we dynamically sample a single MT sample per 
HT sample. As a result, a different class-balanced dataset is obtained for each training epoch.

The multilingual dataset is class-balanced but contains a different number of samples per language, leading to a language-unbalanced distribution. Following \citet{arivazhagan2019massively}, we apply temperature-based sampling, where each language 
$l$ 
is assigned a sampling probability 
proportional to 
$p_l^\frac{1}{T}$, where $p_l$ denotes the original proportion of samples for language $l$. 
Specifically, we use $\frac{1}{T}=0.3$, as suggested by \citet{nllb}. Similarly to the multi-MT dataset, we obtain a different dataset per training epoch.

In contrast to the dynamic generation of training samples, described above, development datasets remain unchanged once created for the multilingual setting. In the multi-MT setting, we build new class-balanced development datasets, where the MT system is randomly sampled in advance, ensuring consistency across experiments.
With this strategy we aim to reduce the risk of model bias due to the sensitivity of the evaluation metric to class imbalance.
To ensure comparability across experiments, the test sets are the same used in the experiments reported in previous sections. 
Appendix \ref{section:appendix-run-variability} reports on the variability across transferability experiments, using new test sets created via the same methodology as the development datasets described above.

\paragraph{Multi-MT Results.} Table~\ref{table:transferability-multiple-mt} presents the results for the multi-MT evaluation, where \ourApproach{} consistently outperforms the baseline by a statistically significant margin in all languages and MT systems, except for 
en--de. This trend also holds in the zero-shot setting with M2M, where en--de is the only exception where the baseline outperforms \ourApproach{}. To assess transferability, we compare these results with those in Table~\ref{table:cross-eval-mt-single-models}. We observe that (i) in most cases, models trained on all available MT systems do not outperform those trained on a single language and MT system (i.e., diagonal values) 
and (ii) both approaches consistently perform better across all languages for cross-MT results (i.e., off-diagonal values), except for de--es and es--de in the baseline (likely due to its English-centric pretraining). These results suggest moderate positive transferability, with minimal accuracy differences on average.

\begin{table*}[tb]
   \centering
   \begin{small}
   \begin{tabular}{cc|rrrrrr||r}
   \toprule
   \multirow{2}{*}{\makecell{Train.\\\& test MT}} & \multirow{2}{*}{\thead{Approach}} & \multicolumn{7}{c}{\thead{Test on lang. data}} \\

   & & de--en& ru--en& en--de& en--ru& de--es& es--de& fi--en \\

   \midrule
   \multirow{2}{*}{Google}
   &  \multirow{1}{*}{\biBsl{}}
&  74.80 &  72.60 &  - &  - &  - &  - &  76.53 \\

   &  \multirow{1}{*}{\ourApproach{}}
&  \textbf{75.52} &  \underline{\textbf{75.85}} &  - &  - &  - &  - &  \underline{\textbf{78.43}} \\

   \midrule
   \multirow{2}{*}{DeepL}
   &  \multirow{1}{*}{\biBsl{}}
&  72.80 &  65.70 &  - &  - &  - &  - &  70.69 \\

   &  \multirow{1}{*}{\ourApproach{}}
&  \textbf{72.83} &  \underline{\textbf{67.35}} &  - &  - &  - &  - &  \underline{\textbf{73.35}} \\

   \midrule
   \multirow{2}{*}{Tower}
   &  \multirow{1}{*}{\biBsl{}}
&  70.92 &  68.27 &  62.19 &  64.27 &  70.57 &  70.47 &  \underline{\textbf{66.66}} \\

   &  \multirow{1}{*}{\ourApproach{}}
&  \textbf{71.47} &  \underline{\textbf{72.22}} &  \textbf{62.79} &  \underline{\textbf{67.70}} &  \underline{\textbf{75.30}} &  \underline{\textbf{76.36}} &  62.83 \\

   \midrule
   \multirow{2}{*}{MADL.}
   &  \multirow{1}{*}{\biBsl{}}
&  - &  - &  69.03 &  70.46 &  71.71 &  71.51 &  66.31 \\

   &  \multirow{1}{*}{\ourApproach{}}
&  - &  - &  \textbf{70.28} &  \underline{\textbf{77.17}} &  \underline{\textbf{82.74}} &  \underline{\textbf{82.25}} &  \underline{\textbf{72.44}} \\

   \midrule
   \multirow{2}{*}{Opus}
   &  \multirow{1}{*}{\biBsl{}}
&  - &  - &  75.21 &  \underline{\textbf{74.16}} &  73.93 &  77.67 &  \textbf{66.03} \\

   &  \multirow{1}{*}{\ourApproach{}}
&  - &  - &  \textbf{75.84} &  71.06 &  \underline{\textbf{77.84}} &  \underline{\textbf{81.97}} &  65.16 \\

   \bottomrule
   \end{tabular}
   \end{small}
   
   \caption{Accuracy obtained by the bilingual baseline (LM) and \ourApproach{} in the multilingual setting. Results are provided per language pair (columns) for models trained on all the available language pairs for a given MT system (rows). The language pair fi--en is only used for testing (zero-shot). Best scores across approaches are in bold. Underlined values indicate statistically significant superior performance across systems.}
   \label{table:transferability-multiple-lang}
\end{table*}

\paragraph{Multilingual Results.} Table~\ref{table:transferability-multiple-lang} presents the results on the multilingual evaluation.
While the results for \ourApproach{} and the baseline are not consistent across MT systems, \ourApproach{} outperforms the baseline in most cases for different languages per MT system, except for some mixed results with Opus;\footnote{Opus is a bilingual MT system, where each language pair corresponds to a different MT model, unlike the other MT systems we evaluate. It is unclear whether Google and DeepL operate as bilingual systems.} 
this trend also holds for nearly all results in the zero-shot language pair (fi--en).
In contrast to models trained on multiple MT systems above, models trained on multiple languages per MT system generally outperform those trained on a single language and MT system (see diagonal values in Table~\ref{table:cross-eval-lang-single-models}). 
These results suggest that incorporating training data from multiple languages improves transferability for both approaches, enabling a single model to operate across multiple languages, albeit for a single MT system.

The results above raise the question whether training a model on both multiple languages and multiple MT systems could further improve performance. Experiments conducted combining data from the multilingual and multi-MT datasets show that, consistent with our earlier findings, training on multiple languages improves performance, but incorporating translations from multiple MT systems degrades performance. In any case, a model trained on all available data achieves competitive results compared to models trained on a single language and system, offering the advantage of broader coverage with a single model.\footnote{Results not provided due to space constraints.}

\begin{table*}[t!]
    \centering
    \begin{small}
    
    \begin{tabular}{@{\hspace{1ex}}c@{\hspace{2ex}}l||@{\hspace{1ex}}r@{\hspace{1.5ex}}r@{\hspace{1.5ex}}r@{\hspace{1.5ex}}r@{\hspace{1ex}}|@{\hspace{1ex}}r@{\hspace{1.5ex}}r@{\hspace{1.5ex}}r@{\hspace{1.5ex}}r@{\hspace{1ex}}|@{\hspace{1ex}}r@{\hspace{1.5ex}}r@{\hspace{1.5ex}}r@{\hspace{1.5ex}}r@{\hspace{1ex}}}
    \toprule
    \multirow{4}{*}{\makecell{Train.\\\& eval.\\lang.}} & \multirow{4}{*}{\thead{Approach}} & \multicolumn{12}{c}{\multirow{2}{*}{\thead{Train. \& eval. MT system}}} \\
    & & \multicolumn{3}{c}{} \\
    & & \multicolumn{4}{c}{Google}& \multicolumn{4}{c}{DeepL} & \multicolumn{4}{c}{Tower} \\
    
    & & Min. & Mean & Max. & SD & Min. & Mean & Max. & SD & Min. & Mean & Max. & SD \\

    \midrule
    \multirow{5}{*}{de--en}

    & LLMixtic & 49.53 & 50.20 & 50.55 & 0.58 & 49.18 & 49.65 & 49.98 & 0.42 & 49.43 & 53.42 & 56.70 & 3.69 \\
    & \monoBsl{} & 63.40 & 63.81 & 64.07 & 0.36 & 61.95 & 62.72 & 63.13 & 0.67 & 61.10 & 61.34 & 61.48 & 0.21 \\
    & \biBsl{} & 75.15 & 75.44 & 75.65 & 0.26 & 71.95 & 72.17 & 72.53 & 0.31 & 69.60 & 69.94 & 70.38 & 0.40 \\
    & \ourApproach{} & 74.80 & 75.44 & 75.88 & 0.57 & 72.90 & 73.48 & 74.00 & 0.55 & 71.00 & 71.27 & 71.47 & 0.24 \\
    & \ourApproachCombined{} & 77.28 & 77.64 & 77.93 & 0.33 & 74.15 & 74.59 & 75.25 & 0.58 & 71.80 & 72.21 & 72.50 & 0.36 \\

    \midrule
    \multirow{5}{*}{ru--en}

    & LLMixtic & 48.90 & 49.19 & 49.45 & 0.28 & 50.38 & 50.57 & 50.68 & 0.17 & 49.25 & 49.38 & 49.60 & 0.20 \\
    & \monoBsl{} & 58.25 & 58.97 & 59.53 & 0.65 & 57.17 & 58.00 & 59.58 & 1.36 & 51.82 & 54.16 & 56.60 & 2.39 \\
    & \biBsl{} & 68.30 & 69.79 & 70.75 & 1.31 & 62.75 & 63.73 & 64.38 & 0.86 & 62.60 & 63.17 & 64.07 & 0.79 \\
    & \ourApproach{} & 73.43 & 73.89 & 74.40 & 0.49 & 64.75 & 64.98 & 65.40 & 0.36 & 66.62 & 67.40 & 68.17 & 0.78 \\
    & \ourApproachCombined{} & 71.72 & 72.92 & 74.08 & 1.18 & 64.90 & 64.98 & 65.10 & 0.10 & 65.97 & 66.70 & 67.42 & 0.73 \\

    \midrule

    & & \multicolumn{4}{c}{MADLAD}& \multicolumn{4}{c}{Opus} & \multicolumn{4}{c}{Tower} \\

    \midrule
    \multirow{5}{*}{en--de}

    & LLMixtic & 49.97 & 50.78 & 52.35 & 1.37 & 50.00 & 50.39 & 50.70 & 0.36 & 53.71 & 53.94 & 54.18 & 0.24 \\
    & \monoBsl{} & 58.96 & 59.20 & 59.49 & 0.27 & 64.55 & 64.85 & 65.20 & 0.33 & 55.18 & 56.19 & 57.31 & 1.07 \\
    & \biBsl{} & 70.23 & 70.45 & 70.76 & 0.27 & 73.69 & 73.99 & 74.39 & 0.36 & 60.72 & 60.92 & 61.04 & 0.18 \\
    & \ourApproach{} & 72.26 & 72.98 & 73.71 & 0.73 & 75.34 & 75.76 & 76.09 & 0.39 & 51.48 & 57.41 & 60.62 & 5.14 \\
    & \ourApproachCombined{} & 73.69 & 74.04 & 74.24 & 0.31 & 77.92 & 78.22 & 78.54 & 0.31 & 60.27 & 60.94 & 62.27 & 1.15 \\

    \midrule
    \multirow{5}{*}{en--ru}

    & LLMixtic & 49.62 & 50.27 & 51.18 & 0.81 & 50.50 & 52.49 & 53.98 & 1.79 & 51.75 & 51.83 & 51.90 & 0.08 \\
    & \monoBsl{} & 56.61 & 61.37 & 63.97 & 4.13 & 68.00 & 69.09 & 69.78 & 0.95 & 57.06 & 57.79 & 58.76 & 0.88 \\
    & \biBsl{} & 61.79 & 66.25 & 68.70 & 3.87 & 70.38 & 70.91 & 71.38 & 0.50 & 60.77 & 60.97 & 61.37 & 0.34 \\
    & \ourApproach{} & 71.83 & 73.11 & 73.99 & 1.13 & 68.60 & 69.80 & 71.48 & 1.50 & 62.59 & 63.40 & 64.17 & 0.79 \\
    & \ourApproachCombined{} & 74.06 & 75.54 & 76.46 & 1.29 & 69.38 & 71.74 & 73.41 & 2.10 & 64.45 & 64.61 & 64.77 & 0.16 \\

    \midrule
    \multirow{5}{*}{de--es}

    & LLMixtic & 55.86 & 56.51 & 57.18 & 0.66 & 51.00 & 51.38 & 51.63 & 0.33 & 53.13 & 54.81 & 55.86 & 1.47 \\
    & \monoBsl{} & 56.72 & 57.55 & 58.31 & 0.80 & 55.95 & 61.11 & 66.26 & 5.15 & 60.08 & 60.79 & 61.67 & 0.81 \\
    & \biBsl{} & 63.03 & 64.37 & 66.35 & 1.75 & 65.40 & 65.85 & 66.39 & 0.51 & 63.35 & 63.58 & 63.90 & 0.28 \\
    & \ourApproach{} & 76.52 & 78.29 & 80.20 & 1.84 & 74.98 & 75.58 & 76.66 & 0.93 & 51.00 & 64.65 & 71.71 & 11.83 \\
    & \ourApproachCombined{} & 73.43 & 75.61 & 77.34 & 1.99 & 73.66 & 74.69 & 76.07 & 1.24 & 64.67 & 67.35 & 70.48 & 2.93 \\

    \midrule
    \multirow{5}{*}{es--de}

    & LLMixtic & 53.62 & 54.47 & 55.53 & 0.97 & 50.09 & 50.39 & 50.68 & 0.29 & 50.45 & 50.79 & 51.09 & 0.32 \\
    & \monoBsl{} & 57.84 & 58.62 & 59.10 & 0.69 & 62.64 & 64.13 & 66.17 & 1.83 & 52.85 & 54.15 & 55.30 & 1.23 \\
    & \biBsl{} & 63.59 & 65.08 & 67.30 & 1.96 & 63.68 & 64.52 & 65.81 & 1.13 & 59.78 & 60.98 & 61.68 & 1.04 \\
    & \ourApproach{} & 74.23 & 75.75 & 76.95 & 1.39 & 75.00 & 76.39 & 77.31 & 1.22 & 69.16 & 70.73 & 72.46 & 1.66 \\
    & \ourApproachCombined{} & 69.11 & 73.32 & 78.03 & 4.48 & 73.55 & 75.23 & 78.12 & 2.52 & 61.37 & 62.58 & 64.31 & 1.54 \\

    \bottomrule
    \end{tabular}
    \end{small}
    \caption{
    Variability over the test set for the experiments reported in Sec.~\ref{ssection:main-results}. Note that the values in the column reporting the maximum do not always match those in Table~\ref{table:results-baseline-vs-ours-vs-frozen}, as those correspond to the training execution achieving the highest accuracy on the development set (see Sec.~\ref{section:results}). SD refers to standard deviation.
    }
    \label{table:variability-main-results}
\end{table*}

\section{Related Work}
\label{section:related-work}

Several studies have explored differences between human and machine-generated translations \citep{vanmassenhove2019lost,roberts2020decoding,luo2024diverge}. However, modern MT systems now produce highly convincing texts, making the HT vs.\ MT classification task increasingly challenging ---evidence suggests that higher-quality MT outputs are harder to detect \citep{aharoni2014automatic}.

Early approaches focused on identifying statistical MT outputs by leveraging fluency and linguistic features \citep{arase2013machine,li2015machine}. \citet{nguyen2017detecting} was the first to address neural MT detection, focusing on distinguishing MT from original texts (rather than human translations) using n-gram-based fluency and noise features. Most studies have concentrated on sentence-level classification, primarily motivated by the fact that neural MT systems are typically trained at this level, despite emerging trends toward coarser granularity \citep{kocmi2024findings}. \citet{bhardwaj2020human} used feature-based models, recurrent neural networks, and pre-trained monolingual and multilingual transformers across several English--French domains. 
\citet{fu2021automatic} analyzed lexical diversity via n-grams and BERT models for English translations from multiple source languages. \citet{nguyen2021machine} proposed an alternative approach measuring differences in machine-generated texts iteratively back-translating them.

Given that classification becomes more challenging for shorter texts \citep{bhardwaj2020human,nguyen2021machine}, several studies have explored coarser granularities beyond the sentence level. \citet{nguyen2017identifying} leveraged Zipfian distributions at the document level, while \citet{nguyen2018identifying,nguyen2019detecting} assessed paragraph coherence. \citet{van2022automatic} found that document-level evaluation outperforms sentence-level detection for German--English texts, using SVMs and various pre-trained monolingual transformer-based classifiers. Building on this, \citet{chichirau2023automatic} extended the analysis to seven source languages for English translations, incorporating multilingual models.

Concurrent to our work, 
\citet{chen2025potential} propose combining a surrogate speech model with a monolingual encoder-based LM to differentiate original, human-produced (non-translated) text from MT-generated text. Our approach differs, not only in the addressed task, but also in that we use a neural MT surrogate model and leverage its internal representations without fine-tuning, making it broadly applicable across the languages supported by the surrogate. 
Moreover, their motivation lies in leveraging speech features to better capture the linguistic nuances and deviations characteristic of machine-generated text, whereas our approach is motivated by modeling the underlying probability distributions of human and machine-generated translations directly, using a surrogate model trained for MT ---which is more aligned with the HT vs.\ MT classification task.

\begin{table}[t!]
    \centering
    \begin{small}
    
    \begin{tabular}{@{\hspace{1ex}}c@{\hspace{2ex}}l|@{\hspace{1ex}}r@{\hspace{1.5ex}}r@{\hspace{1.5ex}}r@{\hspace{1.5ex}}r@{\hspace{1ex}}}
    \toprule
    \multirow{2}{*}{\makecell{Train. \&\\eval. lang.}} & \multirow{2}{*}{\thead{Approach}} & \multicolumn{4}{c}{\thead{Eval. on all MT systems}} \\
    
    & & Min. & Mean & Max. & SD \\

    \midrule
    \multirow{2}{*}{de--en}
    & \biBsl{} & 69.73 & 69.86 & 69.92 & 0.12 \\
    & \ourApproach{} & 72.87 & 73.19 & 73.65 & 0.41 \\

    \midrule

    \multirow{2}{*}{ru--en}
    & \biBsl{} & 61.83 & 62.60 & 63.17 & 0.70 \\
    & \ourApproach{} & 65.97 & 66.72 & 67.35 & 0.70 \\

    \midrule

    \multirow{2}{*}{en--de}
    & \biBsl{} & 64.87 & 66.36 & 67.25 & 1.30 \\
    & \ourApproach{} & 62.97 & 64.15 & 65.35 & 1.19 \\

    \midrule

    \multirow{2}{*}{en--ru}
    & \biBsl{} & 59.54 & 61.79 & 63.85 & 2.16 \\
    & \ourApproach{} & 64.70 & 67.52 & 70.28 & 2.79 \\

    \midrule

    \multirow{2}{*}{de--es}
    & \biBsl{} & 59.99 & 61.25 & 62.53 & 1.27 \\
    & \ourApproach{} & 50.00 & 65.18 & 73.12 & 13.15 \\

    \midrule

    \multirow{2}{*}{es--de}
    & \biBsl{} & 59.74 & 60.63 & 61.96 & 1.17 \\
    & \ourApproach{} & 72.74 & 73.49 & 74.32 & 0.80 \\

    \bottomrule
    \end{tabular}
    \end{small}
    \caption{
    Variability over the multi-MT test set described in Sec.~\ref{ssection:transferability-assessment} and the corresponding multi-MT experiments presented in that section. SD refers to standard deviation.
    }
    \label{table:variability-transferability-cross-mt}
\end{table}

\section{Concluding Remarks}
\label{section:concluding-remarks}

We have presented an 
approach, \ourApproach{}, that directly leverages the internal representations of a surrogate neural MT model to address the HT vs.\ MT classification task. Additionally, we have also introduced \ourApproachCombined{}, an extension that integrates an encoder-based language model.

Results across different surrogate model sizes indicate that larger models achieve better performance, with particularly strong results when using representations from middle decoder blocks. Importantly, we improve upon the current state of the art for this task, with notable gains ---exceeding 5 accuracy points--- for language pairs that do not involve English. By leveraging representations from a multilingual surrogate model that is not primarily focused on English, our approach effectively addresses non-English language pairs.

Cross-evaluation experiments show strong generalization capabilities, while transferability experiments indicate improvements when transferring across languages, but some degradation when transferring across MT systems. 

Finally, the strength of our approach is further validated in a zero-shot setting, where our method achieves gains of 10--15 accuracy points on non-English pairs using an unseen MT system and language.

Code, datasets, and models are available at \url{https://github.com/transducens/SMaTD}.

\begin{table}[t!]
    \centering
    \begin{small}
    
    \begin{tabular}{@{\hspace{1ex}}c@{\hspace{2ex}}l|@{\hspace{1ex}}r@{\hspace{1.5ex}}r@{\hspace{1.5ex}}r@{\hspace{1.5ex}}r@{\hspace{1ex}}}
    \toprule
    \multirow{2}{*}{\makecell{Train. \&\\eval. MT}} & \multirow{2}{*}{\thead{Approach}} & \multicolumn{4}{c}{\thead{Eval. on all lang. data}} \\
    
    & & Min. & Mean & Max. & SD \\

    \midrule
    \multirow{2}{*}{Google}
    & \biBsl{} & 72.37 & 73.05 & 73.68 & 0.66 \\
    & \ourApproach{} & 74.06 & 74.15 & 74.26 & 0.10 \\

    \midrule

    \multirow{2}{*}{DeepL}
    & \biBsl{} & 70.63 & 71.46 & 72.27 & 0.82 \\
    & \ourApproach{} & 71.74 & 72.08 & 72.31 & 0.30 \\

    \midrule

    \multirow{2}{*}{Tower}
    & \biBsl{} & 67.27 & 67.60 & 68.20 & 0.53 \\
    & \ourApproach{} & 69.10 & 69.44 & 69.86 & 0.39 \\

    \midrule

    \multirow{2}{*}{MADL.}
    & \biBsl{} & 72.97 & 73.73 & 74.25 & 0.68 \\
    & \ourApproach{} & 78.72 & 79.10 & 79.64 & 0.49 \\

    \midrule

    \multirow{2}{*}{Opus}
    & \biBsl{} & 77.14 & 77.79 & 78.51 & 0.69 \\
    & \ourApproach{} & 78.23 & 78.44 & 78.64 & 0.21 \\

    \bottomrule
    \end{tabular}
    \end{small}
    \caption{
    Variability over the multilingual test set described in Sec.~\ref{ssection:transferability-assessment} and the corresponding multilingual experiments presented in that section. SD refers to standard deviation.
    }
    \label{table:variability-transferability-cross-lingual}
\end{table}

\section{Limitations}

Our analysis is limited to the news domain, and we have not evaluated the robustness of our approach across domains. We also relied on a single surrogate system, without exploring alternative encoder-decoder or decoder-only architectures, or how specific components affect HT vs.\ MT classification. Additionally, the MT outputs used in our experiments were primarily generated by 
state-of-the-art 
encoder-decoder models with standard decoding strategies, leaving the effects of alternative decoding methods 
---as well as other paradigms, such as statistical or rule-based systems, and architectures like decoder-only transformers or recurrent neural networks--- 
underexplored. Finally, our work does not address translation detection for systems trained at coarser granularities, such as document-level translation.

\appendix

\section{Variability Across Training Executions}
\label{section:appendix-run-variability}

Table~\ref{table:variability-main-results} reports the variability of the accuracy computed on the test set across three training executions with different random seeds for the main experiments (Sec.~\ref{ssection:main-results}). Variability for the transferability experiments (Sec.~\ref{ssection:transferability-assessment}) are 
presented in Table~\ref{table:variability-transferability-cross-mt} and Table~\ref{table:variability-transferability-cross-lingual} for the multi-MT and multilingual experiments, respectively.

\section*{Acknowledgments}
We thank the action editor and anonymous reviewers of the manuscript for their valuable comments which helped us improve the paper.

This paper is part of the work conducted in R+D+i projects PID2021-27999NB-I00 and PID2024-158157OB-C31 funded by the Spanish Ministry of Science and Innovation (MCIN), the Spanish Research Agency (AEI/10.13039/501100011033) and the European Regional Development Fund A way to make Europe. 
Cristian García-Romero is funded by Generalitat Valenciana and the European Social Fund through the research grant CIACIF/2021/365. 
Some of the computational resources used were funded by the Valencia Government and the European Regional Development Fund (ERDF) through project IDIFEDER/2020/003.

\bibliography{main}

\begin{thebibliography}{43}
\expandafter\ifx\csname natexlab\endcsname\relax\def\natexlab#1{#1}\fi

\bibitem[{Aharoni et~al.(2014)Aharoni, Koppel, and Goldberg}]{aharoni2014automatic}
Roee Aharoni, Moshe Koppel, and Yoav Goldberg. 2014.
\newblock \href {https://doi.org/10.3115/v1/P14-2048} {Automatic detection of machine translated text and translation quality estimation}.
\newblock In \emph{Proceedings of the 52nd Annual Meeting of the Association for Computational Linguistics (Volume 2: Short Papers)}, pages 289--295, Baltimore, Maryland. Association for Computational Linguistics.

\bibitem[{Alves et~al.(2024)Alves, Pombal, Guerreiro, Martins, Alves, Farajian, Peters, Rei, Fernandes, Agrawal, Colombo, de~Souza, and Martins}]{alves2024tower}
Duarte~M. Alves, Jos{\'e} Pombal, Nuno~M. Guerreiro, Pedro~H. Martins, Jo{\~a}o Alves, Amin Farajian, Ben Peters, Ricardo Rei, Patrick Fernandes, Sweta Agrawal, Pierre Colombo, Jos{\'e} G.~C. de~Souza, and Andre Martins. 2024.
\newblock \href {https://openreview.net/forum?id=EHPns3hVkj} {Tower: An open multilingual large language model for translation-related tasks}.
\newblock In \emph{First Conference on Language Modeling}.

\bibitem[{Arase and Zhou(2013)}]{arase2013machine}
Yuki Arase and Ming Zhou. 2013.
\newblock \href {https://aclanthology.org/P13-1157/} {Machine translation detection from monolingual web-text}.
\newblock In \emph{Proceedings of the 51st Annual Meeting of the Association for Computational Linguistics (Volume 1: Long Papers)}, pages 1597--1607, Sofia, Bulgaria. Association for Computational Linguistics.

\bibitem[{Arivazhagan et~al.(2019)Arivazhagan, Bapna, Firat, Lepikhin, Johnson, Krikun, Chen, Cao, Foster, Cherry, Macherey, Chen, and Wu}]{arivazhagan2019massively}
Naveen Arivazhagan, Ankur Bapna, Orhan Firat, Dmitry Lepikhin, Melvin Johnson, Maxim Krikun, Mia~Xu Chen, Yuan Cao, George Foster, Colin Cherry, Wolfgang Macherey, Zhifeng Chen, and Yonghui Wu. 2019.
\newblock \href {https://doi.org/10.48550/arXiv.1907.05019} {Massively multilingual neural machine translation in the wild: Findings and challenges}.
\newblock \emph{arXiv preprint arXiv:1907.05019}.

\bibitem[{Bhardwaj et~al.(2020)Bhardwaj, Alfonso~Hermelo, Langlais, Bernier-Colborne, Goutte, and Simard}]{bhardwaj2020human}
Shivendra Bhardwaj, David Alfonso~Hermelo, Phillippe Langlais, Gabriel Bernier-Colborne, Cyril Goutte, and Michel Simard. 2020.
\newblock \href {https://doi.org/10.18653/v1/2020.coling-main.576} {Human or neural translation?}
\newblock In \emph{Proceedings of the 28th International Conference on Computational Linguistics}, pages 6553--6564, Barcelona, Spain (Online). International Committee on Computational Linguistics.

\bibitem[{Chen et~al.(2025)Chen, Farr{\'u}s, and Toral}]{chen2025potential}
Yongjian Chen, Mireia Farr{\'u}s, and Antonio Toral. 2025.
\newblock \href {https://doi.org/10.1109/ICASSP49660.2025.10887578} {The potential of speech features to discriminate between original and machine-translated texts}.
\newblock In \emph{ICASSP 2025-2025 IEEE International Conference on Acoustics, Speech and Signal Processing (ICASSP)}, pages 1--5. IEEE.

\bibitem[{Chichirau et~al.(2023)Chichirau, van Noord, and Toral}]{chichirau2023automatic}
Malina Chichirau, Rik van Noord, and Antonio Toral. 2023.
\newblock \href {https://aclanthology.org/2023.eamt-1.21/} {Automatic discrimination of human and neural machine translation in multilingual scenarios}.
\newblock In \emph{Proceedings of the 24th Annual Conference of the European Association for Machine Translation}, pages 217--226, Tampere, Finland. European Association for Machine Translation.

\bibitem[{Conneau et~al.(2020)Conneau, Khandelwal, Goyal, Chaudhary, Wenzek, Guzm{\'a}n, Grave, Ott, Zettlemoyer, and Stoyanov}]{conneau2020unsupervised}
Alexis Conneau, Kartikay Khandelwal, Naman Goyal, Vishrav Chaudhary, Guillaume Wenzek, Francisco Guzm{\'a}n, Edouard Grave, Myle Ott, Luke Zettlemoyer, and Veselin Stoyanov. 2020.
\newblock \href {https://doi.org/10.18653/v1/2020.acl-main.747} {Unsupervised cross-lingual representation learning at scale}.
\newblock In \emph{Proceedings of the 58th Annual Meeting of the Association for Computational Linguistics}, pages 8440--8451, Online. Association for Computational Linguistics.

\bibitem[{Devlin et~al.(2019)Devlin, Chang, Lee, and Toutanova}]{devlin-etal-2019-bert}
Jacob Devlin, Ming-Wei Chang, Kenton Lee, and Kristina Toutanova. 2019.
\newblock \href {https://doi.org/10.18653/v1/N19-1423} {{BERT}: Pre-training of deep bidirectional transformers for language understanding}.
\newblock In \emph{Proceedings of the 2019 Conference of the North {A}merican Chapter of the Association for Computational Linguistics: Human Language Technologies, Volume 1 (Long and Short Papers)}, pages 4171--4186, Minneapolis, Minnesota. Association for Computational Linguistics.

\bibitem[{Dodge et~al.(2021)Dodge, Sap, Marasovi{\'c}, Agnew, Ilharco, Groeneveld, Mitchell, and Gardner}]{dodge2021documenting}
Jesse Dodge, Maarten Sap, Ana Marasovi{\'c}, William Agnew, Gabriel Ilharco, Dirk Groeneveld, Margaret Mitchell, and Matt Gardner. 2021.
\newblock \href {https://doi.org/10.18653/v1/2021.emnlp-main.98} {Documenting large webtext corpora: A case study on the colossal clean crawled corpus}.
\newblock In \emph{Proceedings of the 2021 Conference on Empirical Methods in Natural Language Processing}, pages 1286--1305, Online and Punta Cana, Dominican Republic. Association for Computational Linguistics.

\bibitem[{Fan et~al.(2021)Fan, Bhosale, Schwenk, Ma, El-Kishky, Goyal, Baines, Celebi, Wenzek, Chaudhary, Goyal, Birch, Liptchinsky, Edunov, Auli, and Joulin}]{fan2021beyond}
Angela Fan, Shruti Bhosale, Holger Schwenk, Zhiyi Ma, Ahmed El-Kishky, Siddharth Goyal, Mandeep Baines, Onur Celebi, Guillaume Wenzek, Vishrav Chaudhary, Naman Goyal, Tom Birch, Vitaliy Liptchinsky, Sergey Edunov, Michael Auli, and Armand Joulin. 2021.
\newblock \href {http://jmlr.org/papers/v22/20-1307.html} {Beyond english-centric multilingual machine translation}.
\newblock \emph{Journal of Machine Learning Research}, 22(107):1--48.

\bibitem[{Fu and Nederhof(2021)}]{fu2021automatic}
Yingxue Fu and Mark-Jan Nederhof. 2021.
\newblock \href {https://aclanthology.org/2021.motra-1.10/} {Automatic classification of human translation and machine translation: A study from the perspective of lexical diversity}.
\newblock In \emph{Proceedings for the First Workshop on Modelling Translation: Translatology in the Digital Age}, pages 91--99, online. Association for Computational Linguistics.

\bibitem[{Gehrmann et~al.(2019)Gehrmann, Strobelt, and Rush}]{gehrmann2019gltr}
Sebastian Gehrmann, Hendrik Strobelt, and Alexander Rush. 2019.
\newblock \href {https://doi.org/10.18653/v1/P19-3019} {{GLTR}: Statistical detection and visualization of generated text}.
\newblock In \emph{Proceedings of the 57th Annual Meeting of the Association for Computational Linguistics: System Demonstrations}, pages 111--116, Florence, Italy. Association for Computational Linguistics.

\bibitem[{Grattafiori et~al.(2024)Grattafiori, Dubey, Jauhri, Pandey, Kadian, Al-Dahle, Letman, Mathur, Schelten, Vaughan, Yang, Fan, Goyal, Hartshorn, Yang, Mitra, Sravankumar, Korenev, Hinsvark, Rao, Zhang, Rodriguez, Gregerson, Spataru, Roziere, Biron, Tang, Chern, Caucheteux, Nayak, Bi, Marra, McConnell, Keller, Touret, Wu, Wong, Ferrer, Nikolaidis, Allonsius, Song, Pintz, Livshits, Wyatt, Esiobu, Choudhary, Mahajan, Garcia-Olano, Perino, Hupkes, Lakomkin, AlBadawy, Lobanova, Dinan, Smith, Radenovic, Guzmán, Zhang, Synnaeve, Lee, Anderson, Thattai, Nail, Mialon, Pang, Cucurell, Nguyen, Korevaar, Xu, Touvron, Zarov, Ibarra, Kloumann, Misra, Evtimov, Zhang, Copet, Lee, Geffert, Vranes, Park, Mahadeokar, Shah, van~der Linde, Billock, Hong, Lee, Fu, Chi, Huang, Liu, Wang, Yu, Bitton, Spisak, Park, Rocca, Johnstun, Saxe, Jia, Alwala, Prasad, Upasani, Plawiak, Li, Heafield, Stone, El-Arini, Iyer, Malik, Chiu, Bhalla, Lakhotia, Rantala-Yeary, van~der Maaten, Chen, Tan, Jenkins, Martin, Madaan, Malo, Blecher,
  Landzaat, de~Oliveira, Muzzi, Pasupuleti, Singh, Paluri, Kardas, Tsimpoukelli, Oldham, Rita, Pavlova, Kambadur, Lewis, Si, Singh, Hassan, Goyal, Torabi, Bashlykov, Bogoychev, Chatterji, Zhang, Duchenne, Çelebi, Alrassy, Zhang, Li, Vasic, Weng, Bhargava, Dubal, Krishnan, Koura, Xu, He, Dong, Srinivasan, Ganapathy, Calderer, Cabral, Stojnic, Raileanu, Maheswari, Girdhar, Patel, Sauvestre, Polidoro, Sumbaly, Taylor, Silva, Hou, Wang, Hosseini, Chennabasappa, Singh, Bell, Kim, Edunov, Nie, Narang, Raparthy, Shen, Wan, Bhosale, Zhang, Vandenhende, Batra, Whitman, Sootla, Collot, Gururangan, Borodinsky, Herman, Fowler, Sheasha, Georgiou, Scialom, Speckbacher, Mihaylov, Xiao, Karn, Goswami, Gupta, Ramanathan, Kerkez, Gonguet, Do, Vogeti, Albiero, Petrovic, Chu, Xiong, Fu, Meers, Martinet, Wang, Wang, Tan, Xia, Xie, Jia, Wang, Goldschlag, Gaur, Babaei, Wen, Song, Zhang, Li, Mao, Coudert, Yan, Chen, Papakipos, Singh, Srivastava, Jain, Kelsey, Shajnfeld, Gangidi, Victoria, Goldstand, Menon, Sharma, Boesenberg,
  Baevski, Feinstein, Kallet, Sangani, Teo, Yunus, Lupu, Alvarado, Caples, Gu, Ho, Poulton, Ryan, Ramchandani, Dong, Franco, Goyal, Saraf, Chowdhury, Gabriel, Bharambe, Eisenman, Yazdan, James, Maurer, Leonhardi, Huang, Loyd, Paola, Paranjape, Liu, Wu, Ni, Hancock, Wasti, Spence, Stojkovic, Gamido, Montalvo, Parker, Burton, Mejia, Liu, Wang, Kim, Zhou, Hu, Chu, Cai, Tindal, Feichtenhofer, Gao, Civin, Beaty, Kreymer, Li, Adkins, Xu, Testuggine, David, Parikh, Liskovich, Foss, Wang, Le, Holland, Dowling, Jamil, Montgomery, Presani, Hahn, Wood, Le, Brinkman, Arcaute, Dunbar, Smothers, Sun, Kreuk, Tian, Kokkinos, Ozgenel, Caggioni, Kanayet, Seide, Florez, Schwarz, Badeer, Swee, Halpern, Herman, Sizov, Guangyi, Zhang, Lakshminarayanan, Inan, Shojanazeri, Zou, Wang, Zha, Habeeb, Rudolph, Suk, Aspegren, Goldman, Zhan, Damlaj, Molybog, Tufanov, Leontiadis, Veliche, Gat, Weissman, Geboski, Kohli, Lam, Asher, Gaya, Marcus, Tang, Chan, Zhen, Reizenstein, Teboul, Zhong, Jin, Yang, Cummings, Carvill, Shepard, McPhie,
  Torres, Ginsburg, Wang, Wu, U, Saxena, Khandelwal, Zand, Matosich, Veeraraghavan, Michelena, Li, Jagadeesh, Huang, Chawla, Huang, Chen, Garg, A, Silva, Bell, Zhang, Guo, Yu, Moshkovich, Wehrstedt, Khabsa, Avalani, Bhatt, Mankus, Hasson, Lennie, Reso, Groshev, Naumov, Lathi, Keneally, Liu, Seltzer, Valko, Restrepo, Patel, Vyatskov, Samvelyan, Clark, Macey, Wang, Hermoso, Metanat, Rastegari, Bansal, Santhanam, Parks, White, Bawa, Singhal, Egebo, Usunier, Mehta, Laptev, Dong, Cheng, Chernoguz, Hart, Salpekar, Kalinli, Kent, Parekh, Saab, Balaji, Rittner, Bontrager, Roux, Dollar, Zvyagina, Ratanchandani, Yuvraj, Liang, Alao, Rodriguez, Ayub, Murthy, Nayani, Mitra, Parthasarathy, Li, Hogan, Battey, Wang, Howes, Rinott, Mehta, Siby, Bondu, Datta, Chugh, Hunt, Dhillon, Sidorov, Pan, Mahajan, Verma, Yamamoto, Ramaswamy, Lindsay, Lindsay, Feng, Lin, Zha, Patil, Shankar, Zhang, Zhang, Wang, Agarwal, Sajuyigbe, Chintala, Max, Chen, Kehoe, Satterfield, Govindaprasad, Gupta, Deng, Cho, Virk, Subramanian, Choudhury,
  Goldman, Remez, Glaser, Best, Koehler, Robinson, Li, Zhang, Matthews, Chou, Shaked, Vontimitta, Ajayi, Montanez, Mohan, Kumar, Mangla, Ionescu, Poenaru, Mihailescu, Ivanov, Li, Wang, Jiang, Bouaziz, Constable, Tang, Wu, Wang, Wu, Gao, Kleinman, Chen, Hu, Jia, Qi, Li, Zhang, Zhang, Adi, Nam, Yu, Wang, Zhao, Hao, Qian, Li, He, Rait, DeVito, Rosnbrick, Wen, Yang, Zhao, and Ma}]{grattafiori2024llama3herdmodels}
Aaron Grattafiori, Abhimanyu Dubey, Abhinav Jauhri, Abhinav Pandey, Abhishek Kadian, Ahmad Al-Dahle, Aiesha Letman, Akhil Mathur, Alan Schelten, Alex Vaughan, Amy Yang, Angela Fan, Anirudh Goyal, Anthony Hartshorn, Aobo Yang, Archi Mitra, Archie Sravankumar, Artem Korenev, Arthur Hinsvark, Arun Rao, Aston Zhang, Aurelien Rodriguez, Austen Gregerson, Ava Spataru, Baptiste Roziere, Bethany Biron, Binh Tang, Bobbie Chern, Charlotte Caucheteux, Chaya Nayak, Chloe Bi, Chris Marra, Chris McConnell, Christian Keller, Christophe Touret, Chunyang Wu, Corinne Wong, Cristian~Canton Ferrer, Cyrus Nikolaidis, Damien Allonsius, Daniel Song, Danielle Pintz, Danny Livshits, Danny Wyatt, David Esiobu, Dhruv Choudhary, Dhruv Mahajan, Diego Garcia-Olano, Diego Perino, Dieuwke Hupkes, Egor Lakomkin, Ehab AlBadawy, Elina Lobanova, Emily Dinan, Eric~Michael Smith, Filip Radenovic, Francisco Guzmán, Frank Zhang, Gabriel Synnaeve, Gabrielle Lee, Georgia~Lewis Anderson, Govind Thattai, Graeme Nail, Gregoire Mialon, Guan Pang,
  Guillem Cucurell, Hailey Nguyen, Hannah Korevaar, Hu~Xu, Hugo Touvron, Iliyan Zarov, Imanol~Arrieta Ibarra, Isabel Kloumann, Ishan Misra, Ivan Evtimov, Jack Zhang, Jade Copet, Jaewon Lee, Jan Geffert, Jana Vranes, Jason Park, Jay Mahadeokar, Jeet Shah, Jelmer van~der Linde, Jennifer Billock, Jenny Hong, Jenya Lee, Jeremy Fu, Jianfeng Chi, Jianyu Huang, Jiawen Liu, Jie Wang, Jiecao Yu, Joanna Bitton, Joe Spisak, Jongsoo Park, Joseph Rocca, Joshua Johnstun, Joshua Saxe, Junteng Jia, Kalyan~Vasuden Alwala, Karthik Prasad, Kartikeya Upasani, Kate Plawiak, Ke~Li, Kenneth Heafield, Kevin Stone, Khalid El-Arini, Krithika Iyer, Kshitiz Malik, Kuenley Chiu, Kunal Bhalla, Kushal Lakhotia, Lauren Rantala-Yeary, Laurens van~der Maaten, Lawrence Chen, Liang Tan, Liz Jenkins, Louis Martin, Lovish Madaan, Lubo Malo, Lukas Blecher, Lukas Landzaat, Luke de~Oliveira, Madeline Muzzi, Mahesh Pasupuleti, Mannat Singh, Manohar Paluri, Marcin Kardas, Maria Tsimpoukelli, Mathew Oldham, Mathieu Rita, Maya Pavlova, Melanie Kambadur,
  Mike Lewis, Min Si, Mitesh~Kumar Singh, Mona Hassan, Naman Goyal, Narjes Torabi, Nikolay Bashlykov, Nikolay Bogoychev, Niladri Chatterji, Ning Zhang, Olivier Duchenne, Onur Çelebi, Patrick Alrassy, Pengchuan Zhang, Pengwei Li, Petar Vasic, Peter Weng, Prajjwal Bhargava, Pratik Dubal, Praveen Krishnan, Punit~Singh Koura, Puxin Xu, Qing He, Qingxiao Dong, Ragavan Srinivasan, Raj Ganapathy, Ramon Calderer, Ricardo~Silveira Cabral, Robert Stojnic, Roberta Raileanu, Rohan Maheswari, Rohit Girdhar, Rohit Patel, Romain Sauvestre, Ronnie Polidoro, Roshan Sumbaly, Ross Taylor, Ruan Silva, Rui Hou, Rui Wang, Saghar Hosseini, Sahana Chennabasappa, Sanjay Singh, Sean Bell, Seohyun~Sonia Kim, Sergey Edunov, Shaoliang Nie, Sharan Narang, Sharath Raparthy, Sheng Shen, Shengye Wan, Shruti Bhosale, Shun Zhang, Simon Vandenhende, Soumya Batra, Spencer Whitman, Sten Sootla, Stephane Collot, Suchin Gururangan, Sydney Borodinsky, Tamar Herman, Tara Fowler, Tarek Sheasha, Thomas Georgiou, Thomas Scialom, Tobias Speckbacher,
  Todor Mihaylov, Tong Xiao, Ujjwal Karn, Vedanuj Goswami, Vibhor Gupta, Vignesh Ramanathan, Viktor Kerkez, Vincent Gonguet, Virginie Do, Vish Vogeti, Vítor Albiero, Vladan Petrovic, Weiwei Chu, Wenhan Xiong, Wenyin Fu, Whitney Meers, Xavier Martinet, Xiaodong Wang, Xiaofang Wang, Xiaoqing~Ellen Tan, Xide Xia, Xinfeng Xie, Xuchao Jia, Xuewei Wang, Yaelle Goldschlag, Yashesh Gaur, Yasmine Babaei, Yi~Wen, Yiwen Song, Yuchen Zhang, Yue Li, Yuning Mao, Zacharie~Delpierre Coudert, Zheng Yan, Zhengxing Chen, Zoe Papakipos, Aaditya Singh, Aayushi Srivastava, Abha Jain, Adam Kelsey, Adam Shajnfeld, Adithya Gangidi, Adolfo Victoria, Ahuva Goldstand, Ajay Menon, Ajay Sharma, Alex Boesenberg, Alexei Baevski, Allie Feinstein, Amanda Kallet, Amit Sangani, Amos Teo, Anam Yunus, Andrei Lupu, Andres Alvarado, Andrew Caples, Andrew Gu, Andrew Ho, Andrew Poulton, Andrew Ryan, Ankit Ramchandani, Annie Dong, Annie Franco, Anuj Goyal, Aparajita Saraf, Arkabandhu Chowdhury, Ashley Gabriel, Ashwin Bharambe, Assaf Eisenman, Azadeh
  Yazdan, Beau James, Ben Maurer, Benjamin Leonhardi, Bernie Huang, Beth Loyd, Beto~De Paola, Bhargavi Paranjape, Bing Liu, Bo~Wu, Boyu Ni, Braden Hancock, Bram Wasti, Brandon Spence, Brani Stojkovic, Brian Gamido, Britt Montalvo, Carl Parker, Carly Burton, Catalina Mejia, Ce~Liu, Changhan Wang, Changkyu Kim, Chao Zhou, Chester Hu, Ching-Hsiang Chu, Chris Cai, Chris Tindal, Christoph Feichtenhofer, Cynthia Gao, Damon Civin, Dana Beaty, Daniel Kreymer, Daniel Li, David Adkins, David Xu, Davide Testuggine, Delia David, Devi Parikh, Diana Liskovich, Didem Foss, Dingkang Wang, Duc Le, Dustin Holland, Edward Dowling, Eissa Jamil, Elaine Montgomery, Eleonora Presani, Emily Hahn, Emily Wood, Eric-Tuan Le, Erik Brinkman, Esteban Arcaute, Evan Dunbar, Evan Smothers, Fei Sun, Felix Kreuk, Feng Tian, Filippos Kokkinos, Firat Ozgenel, Francesco Caggioni, Frank Kanayet, Frank Seide, Gabriela~Medina Florez, Gabriella Schwarz, Gada Badeer, Georgia Swee, Gil Halpern, Grant Herman, Grigory Sizov, Guangyi, Zhang, Guna
  Lakshminarayanan, Hakan Inan, Hamid Shojanazeri, Han Zou, Hannah Wang, Hanwen Zha, Haroun Habeeb, Harrison Rudolph, Helen Suk, Henry Aspegren, Hunter Goldman, Hongyuan Zhan, Ibrahim Damlaj, Igor Molybog, Igor Tufanov, Ilias Leontiadis, Irina-Elena Veliche, Itai Gat, Jake Weissman, James Geboski, James Kohli, Janice Lam, Japhet Asher, Jean-Baptiste Gaya, Jeff Marcus, Jeff Tang, Jennifer Chan, Jenny Zhen, Jeremy Reizenstein, Jeremy Teboul, Jessica Zhong, Jian Jin, Jingyi Yang, Joe Cummings, Jon Carvill, Jon Shepard, Jonathan McPhie, Jonathan Torres, Josh Ginsburg, Junjie Wang, Kai Wu, Kam~Hou U, Karan Saxena, Kartikay Khandelwal, Katayoun Zand, Kathy Matosich, Kaushik Veeraraghavan, Kelly Michelena, Keqian Li, Kiran Jagadeesh, Kun Huang, Kunal Chawla, Kyle Huang, Lailin Chen, Lakshya Garg, Lavender A, Leandro Silva, Lee Bell, Lei Zhang, Liangpeng Guo, Licheng Yu, Liron Moshkovich, Luca Wehrstedt, Madian Khabsa, Manav Avalani, Manish Bhatt, Martynas Mankus, Matan Hasson, Matthew Lennie, Matthias Reso, Maxim
  Groshev, Maxim Naumov, Maya Lathi, Meghan Keneally, Miao Liu, Michael~L. Seltzer, Michal Valko, Michelle Restrepo, Mihir Patel, Mik Vyatskov, Mikayel Samvelyan, Mike Clark, Mike Macey, Mike Wang, Miquel~Jubert Hermoso, Mo~Metanat, Mohammad Rastegari, Munish Bansal, Nandhini Santhanam, Natascha Parks, Natasha White, Navyata Bawa, Nayan Singhal, Nick Egebo, Nicolas Usunier, Nikhil Mehta, Nikolay~Pavlovich Laptev, Ning Dong, Norman Cheng, Oleg Chernoguz, Olivia Hart, Omkar Salpekar, Ozlem Kalinli, Parkin Kent, Parth Parekh, Paul Saab, Pavan Balaji, Pedro Rittner, Philip Bontrager, Pierre Roux, Piotr Dollar, Polina Zvyagina, Prashant Ratanchandani, Pritish Yuvraj, Qian Liang, Rachad Alao, Rachel Rodriguez, Rafi Ayub, Raghotham Murthy, Raghu Nayani, Rahul Mitra, Rangaprabhu Parthasarathy, Raymond Li, Rebekkah Hogan, Robin Battey, Rocky Wang, Russ Howes, Ruty Rinott, Sachin Mehta, Sachin Siby, Sai~Jayesh Bondu, Samyak Datta, Sara Chugh, Sara Hunt, Sargun Dhillon, Sasha Sidorov, Satadru Pan, Saurabh Mahajan,
  Saurabh Verma, Seiji Yamamoto, Sharadh Ramaswamy, Shaun Lindsay, Shaun Lindsay, Sheng Feng, Shenghao Lin, Shengxin~Cindy Zha, Shishir Patil, Shiva Shankar, Shuqiang Zhang, Shuqiang Zhang, Sinong Wang, Sneha Agarwal, Soji Sajuyigbe, Soumith Chintala, Stephanie Max, Stephen Chen, Steve Kehoe, Steve Satterfield, Sudarshan Govindaprasad, Sumit Gupta, Summer Deng, Sungmin Cho, Sunny Virk, Suraj Subramanian, Sy~Choudhury, Sydney Goldman, Tal Remez, Tamar Glaser, Tamara Best, Thilo Koehler, Thomas Robinson, Tianhe Li, Tianjun Zhang, Tim Matthews, Timothy Chou, Tzook Shaked, Varun Vontimitta, Victoria Ajayi, Victoria Montanez, Vijai Mohan, Vinay~Satish Kumar, Vishal Mangla, Vlad Ionescu, Vlad Poenaru, Vlad~Tiberiu Mihailescu, Vladimir Ivanov, Wei Li, Wenchen Wang, Wenwen Jiang, Wes Bouaziz, Will Constable, Xiaocheng Tang, Xiaojian Wu, Xiaolan Wang, Xilun Wu, Xinbo Gao, Yaniv Kleinman, Yanjun Chen, Ye~Hu, Ye~Jia, Ye~Qi, Yenda Li, Yilin Zhang, Ying Zhang, Yossi Adi, Youngjin Nam, Yu, Wang, Yu~Zhao, Yuchen Hao, Yundi
  Qian, Yunlu Li, Yuzi He, Zach Rait, Zachary DeVito, Zef Rosnbrick, Zhaoduo Wen, Zhenyu Yang, Zhiwei Zhao, and Zhiyu Ma. 2024.
\newblock \href {http://arxiv.org/abs/2407.21783} {The llama 3 herd of models}.

\bibitem[{He et~al.(2023)He, Gao, and Chen}]{he2023debertav3}
Pengcheng He, Jianfeng Gao, and Weizhu Chen. 2023.
\newblock \href {https://openreview.net/forum?id=sE7-XhLxHA} {{D}e{BERT}a{V}3: Improving {D}e{BERT}a using {ELECTRA}-style pre-training with gradient-disentangled embedding sharing}.
\newblock In \emph{The Eleventh International Conference on Learning Representations}.

\bibitem[{Huang et~al.(2016)Huang, Sun, Liu, Sedra, and Weinberger}]{huang2016deep}
Gao Huang, Yu~Sun, Zhuang Liu, Daniel Sedra, and Kilian~Q. Weinberger. 2016.
\newblock \href {https://doi.org/10.1007/978-3-319-46493-0_39} {Deep networks with stochastic depth}.
\newblock In \emph{Computer Vision -- ECCV 2016}, pages 646--661. Springer International Publishing.

\bibitem[{Ji et~al.(2024)Ji, Li, Paul, Paavola, Lin, Chen, O'Brien, Luo, Sch{\"u}tze, Tiedemann, and Haddow}]{ji2024emma}
Shaoxiong Ji, Zihao Li, Indraneil Paul, Jaakko Paavola, Peiqin Lin, Pinzhen Chen, Dayy{\'a}n O'Brien, Hengyu Luo, Hinrich Sch{\"u}tze, J{\"o}rg Tiedemann, and Barry Haddow. 2024.
\newblock \href {https://doi.org/10.48550/arXiv.2409.17892} {{EMMA}-500: Enhancing massively multilingual adaptation of large language models}.
\newblock \emph{arXiv preprint arXiv:2409.17892}.

\bibitem[{Jiao et~al.(2021)Jiao, Yang, Sun, and Liu}]{jiao2021alternated}
Rui Jiao, Zonghan Yang, Maosong Sun, and Yang Liu. 2021.
\newblock \href {https://doi.org/10.18653/v1/2021.findings-acl.160} {Alternated training with synthetic and authentic data for neural machine translation}.
\newblock In \emph{Findings of the Association for Computational Linguistics: ACL-IJCNLP 2021}, pages 1828--1834, Online. Association for Computational Linguistics.

\bibitem[{Kocmi et~al.(2024)Kocmi, Avramidis, Bawden, Bojar, Dvorkovich, Federmann, Fishel, Freitag, Gowda, Grundkiewicz, Haddow, Karpinska, Koehn, Marie, Monz, Murray, Nagata, Popel, Popovi{\'c}, Shmatova, Steingr{\'i}msson, and Zouhar}]{kocmi2024findings}
Tom Kocmi, Eleftherios Avramidis, Rachel Bawden, Ond{\v{r}}ej Bojar, Anton Dvorkovich, Christian Federmann, Mark Fishel, Markus Freitag, Thamme Gowda, Roman Grundkiewicz, Barry Haddow, Marzena Karpinska, Philipp Koehn, Benjamin Marie, Christof Monz, Kenton Murray, Masaaki Nagata, Martin Popel, Maja Popovi{\'c}, Mariya Shmatova, Steinth{\'o}r Steingr{\'i}msson, and Vil{\'e}m Zouhar. 2024.
\newblock \href {https://doi.org/10.18653/v1/2024.wmt-1.1} {Findings of the {WMT}24 general machine translation shared task: The {LLM} era is here but {MT} is not solved yet}.
\newblock In \emph{Proceedings of the Ninth Conference on Machine Translation}, pages 1--46, Miami, Florida, USA. Association for Computational Linguistics.

\bibitem[{Kudugunta et~al.(2023)Kudugunta, Caswell, Zhang, Garcia, Xin, Kusupati, Stella, Bapna, and Firat}]{kudugunta2024madlad}
Sneha Kudugunta, Isaac Caswell, Biao Zhang, Xavier Garcia, Derrick Xin, Aditya Kusupati, Romi Stella, Ankur Bapna, and Orhan Firat. 2023.
\newblock \href {https://proceedings.neurips.cc/paper_files/paper/2023/file/d49042a5d49818711c401d34172f9900-Paper-Datasets_and_Benchmarks.pdf} {{MADLAD}-400: A multilingual and document-level large audited dataset}.
\newblock In \emph{Advances in Neural Information Processing Systems}, volume~36, pages 67284--67296. Curran Associates, Inc.

\bibitem[{Li et~al.(2015)Li, Wang, and Zhao}]{li2015machine}
Yitong Li, Rui Wang, and Hai Zhao. 2015.
\newblock \href {https://aclanthology.org/Y15-2041/} {A machine learning method to distinguish machine translation from human translation}.
\newblock In \emph{Proceedings of the 29th Pacific Asia Conference on Language, Information and Computation: Posters}, pages 354--360, Shanghai, China.

\bibitem[{Loshchilov and Hutter(2019)}]{loshchilov2017decoupled}
Ilya Loshchilov and Frank Hutter. 2019.
\newblock \href {https://openreview.net/forum?id=Bkg6RiCqY7} {Decoupled weight decay regularization}.
\newblock In \emph{International Conference on Learning Representations}.

\bibitem[{Luo et~al.(2024)Luo, Cherry, and Foster}]{luo2024diverge}
Jiaming Luo, Colin Cherry, and George Foster. 2024.
\newblock \href {https://doi.org/10.1162/tacl_a_00645} {To diverge or not to diverge: A morphosyntactic perspective on machine translation vs human translation}.
\newblock \emph{Transactions of the Association for Computational Linguistics}, 12:355--371.

\bibitem[{Mitchell et~al.(2023)Mitchell, Lee, Khazatsky, Manning, and Finn}]{mitchell2023detectgpt}
Eric Mitchell, Yoonho Lee, Alexander Khazatsky, Christopher~D. Manning, and Chelsea Finn. 2023.
\newblock \href {https://proceedings.mlr.press/v202/mitchell23a.html} {{D}etect{GPT}: Zero-shot machine-generated text detection using probability curvature}.
\newblock In \emph{Proceedings of the 40th International Conference on Machine Learning}, pages 24950--24962.

\bibitem[{Nguyen-Son and Echizen(2018)}]{nguyen2017detecting}
Hoang-Quoc Nguyen-Son and Isao Echizen. 2018.
\newblock \href {https://doi.org/10.1007/978-981-10-8438-6_23} {Detecting computer-generated text using fluency and noise features}.
\newblock In \emph{International Conference of the Pacific Association for Computational Linguistics}, pages 288--300, Singapore. Springer Singapore.

\bibitem[{Nguyen-Son et~al.(2018)Nguyen-Son, Nguyen, Tieu, Yamagishi, and Echizen}]{nguyen2018identifying}
Hoang-Quoc Nguyen-Son, Huy~H. Nguyen, Ngoc-Dung~T. Tieu, Junichi Yamagishi, and Isao Echizen. 2018.
\newblock \href {https://aclanthology.org/Y18-1056/} {Identifying computer-translated paragraphs using coherence features}.
\newblock In \emph{Proceedings of the 32nd Pacific Asia Conference on Language, Information and Computation}, Hong Kong. Association for Computational Linguistics.

\bibitem[{Nguyen-Son et~al.(2021)Nguyen-Son, Thao, Hidano, Gupta, and Kiyomoto}]{nguyen2021machine}
Hoang-Quoc Nguyen-Son, Tran Thao, Seira Hidano, Ishita Gupta, and Shinsaku Kiyomoto. 2021.
\newblock \href {https://doi.org/10.18653/v1/2021.naacl-main.462} {Machine translated text detection through text similarity with round-trip translation}.
\newblock In \emph{Proceedings of the 2021 Conference of the North American Chapter of the Association for Computational Linguistics: Human Language Technologies}, pages 5792--5797, Online. Association for Computational Linguistics.

\bibitem[{Nguyen-Son et~al.(2019)Nguyen-Son, Thao, Hidano, and Kiyomoto}]{nguyen2019detecting}
Hoang-Quoc Nguyen-Son, Tran~Phuong Thao, Seira Hidano, and Shinsaku Kiyomoto. 2019.
\newblock \href {https://doi.org/10.1007/978-3-031-24337-0_36} {Detecting machine-translated paragraphs by matching similar words}.
\newblock In \emph{International Conference on Computational Linguistics and Intelligent Text Processing}, pages 521--532. Springer.

\bibitem[{Nguyen-Son et~al.(2017)Nguyen-Son, Tieu, Nguyen, Yamagishi, and Zen}]{nguyen2017identifying}
Hoang-Quoc Nguyen-Son, Ngoc-Dung~T. Tieu, Huy~H. Nguyen, Junichi Yamagishi, and Isao~Echi Zen. 2017.
\newblock \href {https://doi.org/10.1109/APSIPA.2017.8282270} {Identifying computer-generated text using statistical analysis}.
\newblock In \emph{2017 Asia-Pacific Signal and Information Processing Association Annual Summit and Conference (APSIPA ASC)}, pages 1504--1511. IEEE.

\bibitem[{{NLLB Team} et~al.(2022){NLLB Team}, Costa-juss{\`a}, Cross, {\c{C}}elebi, Elbayad, Heafield, Heffernan, Kalbassi, Lam, Licht, Maillard, Sun, Wang, Wenzek, Youngblood, Akula, Barrault, Gonzalez~Mejia, Hansanti, Hoffman, Jarrett, Sadagopan, Rowe, Spruit, Tran, Andrews, Ayan, Bhosale, Edunov, Fan, Gao, Goswami, Guzmán, Koehn, Mourachko, Ropers, Saleem, Schwenk, and Wang}]{nllb}
{NLLB Team}, Marta~R. Costa-juss{\`a}, James Cross, Onur {\c{C}}elebi, Maha Elbayad, Kenneth Heafield, Kevin Heffernan, Elahe Kalbassi, Janice Lam, Daniel Licht, Jean Maillard, Anna Sun, Skyler Wang, Guillaume Wenzek, Al~Youngblood, Bapi Akula, Loic Barrault, Gabriel Gonzalez~Mejia, Prangthip Hansanti, John Hoffman, Semarley Jarrett, Kaushik~Ram Sadagopan, Dirk Rowe, Shannon Spruit, Chau Tran, Pierre Andrews, Necip~Fazil Ayan, Shruti Bhosale, Sergey Edunov, Angela Fan, Cynthia Gao, Vedanuj Goswami, Francisco Guzmán, Philipp Koehn, Alexandre Mourachko, Christophe Ropers, Safiyyah Saleem, Holger Schwenk, and Jeff Wang. 2022.
\newblock \href {https://doi.org/10.48550/arXiv.2207.04672} {No language left behind: Scaling human-centered machine translation}.
\newblock \emph{arXiv preprint arXiv:2207.04672}.

\bibitem[{Ram{\'i}rez-S{\'a}nchez et~al.(2022)Ram{\'i}rez-S{\'a}nchez, Ba{\~n}{\'o}n, Zaragoza-Bernabeu, and Ortiz~Rojas}]{ramirez2022human}
Gema Ram{\'i}rez-S{\'a}nchez, Marta Ba{\~n}{\'o}n, Jaume Zaragoza-Bernabeu, and Sergio Ortiz~Rojas. 2022.
\newblock \href {https://doi.org/10.18653/v1/2022.humeval-1.4} {Human evaluation of web-crawled parallel corpora for machine translation}.
\newblock In \emph{Proceedings of the 2nd Workshop on Human Evaluation of NLP Systems (HumEval)}, pages 32--41, Dublin, Ireland. Association for Computational Linguistics.

\bibitem[{Roberts et~al.(2020)Roberts, Liang, Neubig, and Lipton}]{roberts2020decoding}
Nicholas Roberts, Davis Liang, Graham Neubig, and Zachary~C. Lipton. 2020.
\newblock \href {https://drive.google.com/file/d/1crAS9oknszKV6Gssr9W8zyAupZ-i8sg9/view} {Decoding and diversity in machine translation}.
\newblock In \emph{Proceedings of the Resistance AI Workshop at 34th Conference on Neural Information Processing Systems (NeurIPS 2020)}, Vancouver, CA.

\bibitem[{Sarvazyan et~al.(2024)Sarvazyan, Gonz{\'a}lez, and Franco-salvador}]{sarvazyan2024genaios}
Areg~Mikael Sarvazyan, Jos{\'e}~{\'A}ngel Gonz{\'a}lez, and Marc Franco-salvador. 2024.
\newblock \href {https://doi.org/10.18653/v1/2024.semeval-1.17} {Genaios at {S}em{E}val-2024 task 8: Detecting machine-generated text by mixing language model probabilistic features}.
\newblock In \emph{Proceedings of the 18th International Workshop on Semantic Evaluation (SemEval-2024)}, pages 101--107, Mexico City, Mexico. Association for Computational Linguistics.

\bibitem[{Shumailov et~al.(2024)Shumailov, Shumaylov, Zhao, Papernot, Anderson, and Gal}]{shumailov2024ai}
Ilia Shumailov, Zakhar Shumaylov, Yiren Zhao, Nicolas Papernot, Ross Anderson, and Yarin Gal. 2024.
\newblock \href {https://doi.org/10.1038/s41586-024-07566-y} {{AI} models collapse when trained on recursively generated data}.
\newblock \emph{Nature}, 631(8022):755--759.

\bibitem[{Thompson et~al.(2024)Thompson, Dhaliwal, Frisch, Domhan, and Federico}]{thompson2024shocking}
Brian Thompson, Mehak Dhaliwal, Peter Frisch, Tobias Domhan, and Marcello Federico. 2024.
\newblock \href {https://doi.org/10.18653/v1/2024.findings-acl.103} {A shocking amount of the web is machine translated: Insights from multi-way parallelism}.
\newblock In \emph{Findings of the Association for Computational Linguistics: ACL 2024}, pages 1763--1775, Bangkok, Thailand. Association for Computational Linguistics.

\bibitem[{Tiedemann and Thottingal(2020)}]{tiedemann2020opus}
J{\"o}rg Tiedemann and Santhosh Thottingal. 2020.
\newblock \href {https://aclanthology.org/2020.eamt-1.61/} {{OPUS}-{MT} {--} building open translation services for the world}.
\newblock In \emph{Proceedings of the 22nd Annual Conference of the European Association for Machine Translation}, pages 479--480, Lisboa, Portugal. European Association for Machine Translation.

\bibitem[{Touvron et~al.(2023)Touvron, Martin, Stone, Albert, Almahairi, Babaei, Bashlykov, Batra, Bhargava, Bhosale, Bikel, Blecher, Ferrer, Chen, Cucurull, Esiobu, Fernandes, Fu, Fu, Fuller, Gao, Goswami, Goyal, Hartshorn, Hosseini, Hou, Inan, Kardas, Kerkez, Khabsa, Kloumann, Korenev, Koura, Lachaux, Lavril, Lee, Liskovich, Lu, Mao, Martinet, Mihaylov, Mishra, Molybog, Nie, Poulton, Reizenstein, Rungta, Saladi, Schelten, Silva, Smith, Subramanian, Tan, Tang, Taylor, Williams, Kuan, Xu, Yan, Zarov, Zhang, Fan, Kambadur, Narang, Rodriguez, Stojnic, Edunov, and Scialom}]{touvron2023llama2openfoundation}
Hugo Touvron, Louis Martin, Kevin Stone, Peter Albert, Amjad Almahairi, Yasmine Babaei, Nikolay Bashlykov, Soumya Batra, Prajjwal Bhargava, Shruti Bhosale, Dan Bikel, Lukas Blecher, Cristian~Canton Ferrer, Moya Chen, Guillem Cucurull, David Esiobu, Jude Fernandes, Jeremy Fu, Wenyin Fu, Brian Fuller, Cynthia Gao, Vedanuj Goswami, Naman Goyal, Anthony Hartshorn, Saghar Hosseini, Rui Hou, Hakan Inan, Marcin Kardas, Viktor Kerkez, Madian Khabsa, Isabel Kloumann, Artem Korenev, Punit~Singh Koura, Marie-Anne Lachaux, Thibaut Lavril, Jenya Lee, Diana Liskovich, Yinghai Lu, Yuning Mao, Xavier Martinet, Todor Mihaylov, Pushkar Mishra, Igor Molybog, Yixin Nie, Andrew Poulton, Jeremy Reizenstein, Rashi Rungta, Kalyan Saladi, Alan Schelten, Ruan Silva, Eric~Michael Smith, Ranjan Subramanian, Xiaoqing~Ellen Tan, Binh Tang, Ross Taylor, Adina Williams, Jian~Xiang Kuan, Puxin Xu, Zheng Yan, Iliyan Zarov, Yuchen Zhang, Angela Fan, Melanie Kambadur, Sharan Narang, Aurelien Rodriguez, Robert Stojnic, Sergey Edunov, and Thomas
  Scialom. 2023.
\newblock \href {http://arxiv.org/abs/2307.09288} {Llama 2: Open foundation and fine-tuned chat models}.

\bibitem[{Vanmassenhove et~al.(2019)Vanmassenhove, Shterionov, and Way}]{vanmassenhove2019lost}
Eva Vanmassenhove, Dimitar Shterionov, and Andy Way. 2019.
\newblock \href {https://aclanthology.org/W19-6622/} {Lost in translation: Loss and decay of linguistic richness in machine translation}.
\newblock In \emph{Proceedings of Machine Translation Summit XVII: Research Track}, pages 222--232, Dublin, Ireland. European Association for Machine Translation.

\bibitem[{Vaswani et~al.(2017)Vaswani, Shazeer, Parmar, Uszkoreit, Jones, Gomez, Kaiser, and Polosukhin}]{vaswani2017attention}
Ashish Vaswani, Noam Shazeer, Niki Parmar, Jakob Uszkoreit, Llion Jones, Aidan~N Gomez, {\L}ukasz Kaiser, and Illia Polosukhin. 2017.
\newblock \href {https://proceedings.neurips.cc/paper_files/paper/2017/file/3f5ee243547dee91fbd053c1c4a845aa-Paper.pdf} {Attention is all you need}.
\newblock In \emph{Proceedings of the 31st International Conference on Neural Information Processing Systems}, pages 6000--6010, Long Beach, USA.

\bibitem[{Venkatraman et~al.(2024)Venkatraman, Uchendu, and Lee}]{venkatraman2024gpt}
Saranya Venkatraman, Adaku Uchendu, and Dongwon Lee. 2024.
\newblock \href {https://doi.org/10.18653/v1/2024.findings-naacl.8} {{GPT}-who: An information density-based machine-generated text detector}.
\newblock In \emph{Findings of the Association for Computational Linguistics: NAACL 2024}, pages 103--115, Mexico City, Mexico. Association for Computational Linguistics.

\bibitem[{Wang et~al.(2024)Wang, Mansurov, Ivanov, Su, Shelmanov, Tsvigun, Mohammed~Afzal, Mahmoud, Puccetti, and Arnold}]{wang-etal-2024-semeval-2024}
Yuxia Wang, Jonibek Mansurov, Petar Ivanov, Jinyan Su, Artem Shelmanov, Akim Tsvigun, Osama Mohammed~Afzal, Tarek Mahmoud, Giovanni Puccetti, and Thomas Arnold. 2024.
\newblock \href {https://doi.org/10.18653/v1/2024.semeval-1.279} {{S}em{E}val-2024 task 8: Multidomain, multimodel and multilingual machine-generated text detection}.
\newblock In \emph{Proceedings of the 18th International Workshop on Semantic Evaluation (SemEval-2024)}, pages 2057--2079, Mexico City, Mexico. Association for Computational Linguistics.

\bibitem[{van~der Werff et~al.(2022)van~der Werff, van Noord, and Toral}]{van2022automatic}
Tobias van~der Werff, Rik van Noord, and Antonio Toral. 2022.
\newblock \href {https://aclanthology.org/2022.eamt-1.19/} {Automatic discrimination of human and neural machine translation: A study with multiple pre-trained models and longer context}.
\newblock In \emph{Proceedings of the 23rd Annual Conference of the European Association for Machine Translation}, pages 161--170, Ghent, Belgium. European Association for Machine Translation.

\bibitem[{Wu et~al.(2019)Wu, Wang, Xia, Qin, Lai, and Liu}]{wu2019exploiting}
Lijun Wu, Yiren Wang, Yingce Xia, Tao Qin, Jianhuang Lai, and Tie-Yan Liu. 2019.
\newblock \href {https://doi.org/10.18653/v1/D19-1430} {Exploiting monolingual data at scale for neural machine translation}.
\newblock In \emph{Proceedings of the 2019 Conference on Empirical Methods in Natural Language Processing and the 9th International Joint Conference on Natural Language Processing (EMNLP-IJCNLP)}, pages 4207--4216, Hong Kong, China. Association for Computational Linguistics.

\end{thebibliography}
\bibliographystyle{acl_natbib}

\end{document}